%% file: main_app.tex
\documentclass[10pt,twocolumn,letterpaper]{article}

\usepackage[pagenumbers]{cvpr} % To force page numbers, e.g. for an arXiv version

\definecolor{cvprblue}{rgb}{0.21,0.49,0.74}
\usepackage[pagebackref,breaklinks,colorlinks,allcolors=cvprblue]{hyperref}
\usepackage{xcolor}
\definecolor{lgray}{rgb}{0.9, 0.9, 0.9}
\usepackage[utf8]{inputenc}
\DeclareUnicodeCharacter{00E9}{\'{e}}
\usepackage{arydshln}
\usepackage{tabularx}
\usepackage{multirow}
\usepackage{amsmath}
\usepackage{amssymb}
\usepackage{bbm}

\usepackage{color}
\usepackage{graphicx}
\usepackage{xcolor}
\usepackage{colortbl}
\usepackage{tikz}
\usepackage{amsmath}
\usepackage{booktabs}
\usepackage{nameref}
\usepackage{inconsolata}
\usepackage{arydshln}
\usepackage{tcolorbox}
\usepackage[ruled,linesnumbered,algo2e]{algorithm2e}
\usepackage{url}
\def\paperID{12881} % *** Enter the Paper ID here
\def\confName{CVPR}
\def\confYear{2025}

\title{Are Anomaly Scores Telling the Whole Story? A Benchmark for Multilevel Anomaly Detection}

\author{
{\rm Tri Cao\textsuperscript{1}, 
Minh-Huy Trinh\textsuperscript{3, 4}, 
Ailin Deng\textsuperscript{1}, Quoc-Nam Nguyen\textsuperscript{1}, Khoa Duong\textsuperscript{2}},\\
{\rm Ngai-Man Cheung\textsuperscript{2}, 
Bryan Hooi\textsuperscript{1}}\\
    \textsuperscript{\rm 1}\textit{National University of Singapore},
    \textsuperscript{\rm 2}\textit{Singapore University of Technology and Design},\\
    \textsuperscript{\rm 3}\textit{University of Science},
    \textsuperscript{\rm 4}\textit{Vietnam National University, Ho Chi Minh City.}  \\}
\usepackage{multirow}

\begin{document}
\maketitle

\input{sec/0_abstract}    
\input{sec/1_intro}

\input{sec/2_related}
\input{sec/3_multiAD}

\input{sec/4_benchmark}
\input{sec/5_expanalysis}

\input{sec/6_future}
\input{sec/7_conclusion}

{
    \small
    \bibliographystyle{ieeenat_fullname}
    \bibliography{refs}
}

\onecolumn
\appendix
\input{sec/X_apptmp}

% {
%     \small
%     \bibliographystyle{ieeenat_fullname}
%     \bibliography{refs}
% }

\end{document}

%% file: sec/0_abstract.tex
\begin{abstract}
Anomaly detection (AD) is a machine learning task that identifies anomalies by learning patterns from normal training data.
In many real-world scenarios, anomalies vary in severity, from minor anomalies with little risk to severe abnormalities requiring immediate attention. However, existing models primarily operate in a binary setting, and the anomaly scores they produce are usually based on the deviation of data points from normal data, which may not accurately reflect practical severity. In this paper, we address this gap by making three key contributions. First, we propose a novel setting, Multilevel AD (MAD), in which the anomaly score represents the severity of anomalies in real-world applications, and we highlight its diverse applications across various domains. Second, we introduce a novel benchmark, MAD-Bench, that evaluates models not only on their ability to detect anomalies, but also on how effectively their anomaly scores reflect severity. This benchmark incorporates multiple types of baselines and real-world applications involving severity. Finally, we conduct a comprehensive performance analysis on MAD-Bench. We evaluate models on their ability to assign severity-aligned scores, investigate the correspondence between their performance on binary and multilevel detection, and study their robustness. This analysis offers key insights into improving AD models for practical severity alignment. The code framework and datasets used for the benchmark will be made publicly available.
% \ailin{should we add some key insights or findings in the abstract, e.g., LMMs outperform the traditional methods; }
\end{abstract}

% In this paper, we address this gap by making three key contributions. First, we propose a new setting, Multilevel Anomaly Detection (MAD), which anomaly score should represent the severity of anomalies in real-world application. We also highlight the diverse applications of this setting across domains such as one-class novelty detection, anomaly detection in medical imaging, and industrial inspection.
% Finally, we conduct a comprehensive performance analysis using this benchmark. We analyze their ability to assign severity-aligned anomaly scores across applications, explore the correlation between binary and multilevel detection performance, and assess the impact of anomalous feature area on anomaly scores. Additionally, we examine performance variations across severity levels, the effect of expanding the normal class to include light anomalies, and the robustness of models under input corruption or manipulation in MAD settings.  These contributions provide a deeper understanding of how AD models can be improved to align anomaly detection with practical severity assessments. The code framework and datasets used for the benchmark will be made publicly available.

%% file: sec/1_intro.tex
% \vspace{-0.5cm}
\section{Introduction}
\label{sec:intro}
\begin{figure*}[!ht]
    \centering
    \includegraphics[width=1\textwidth]{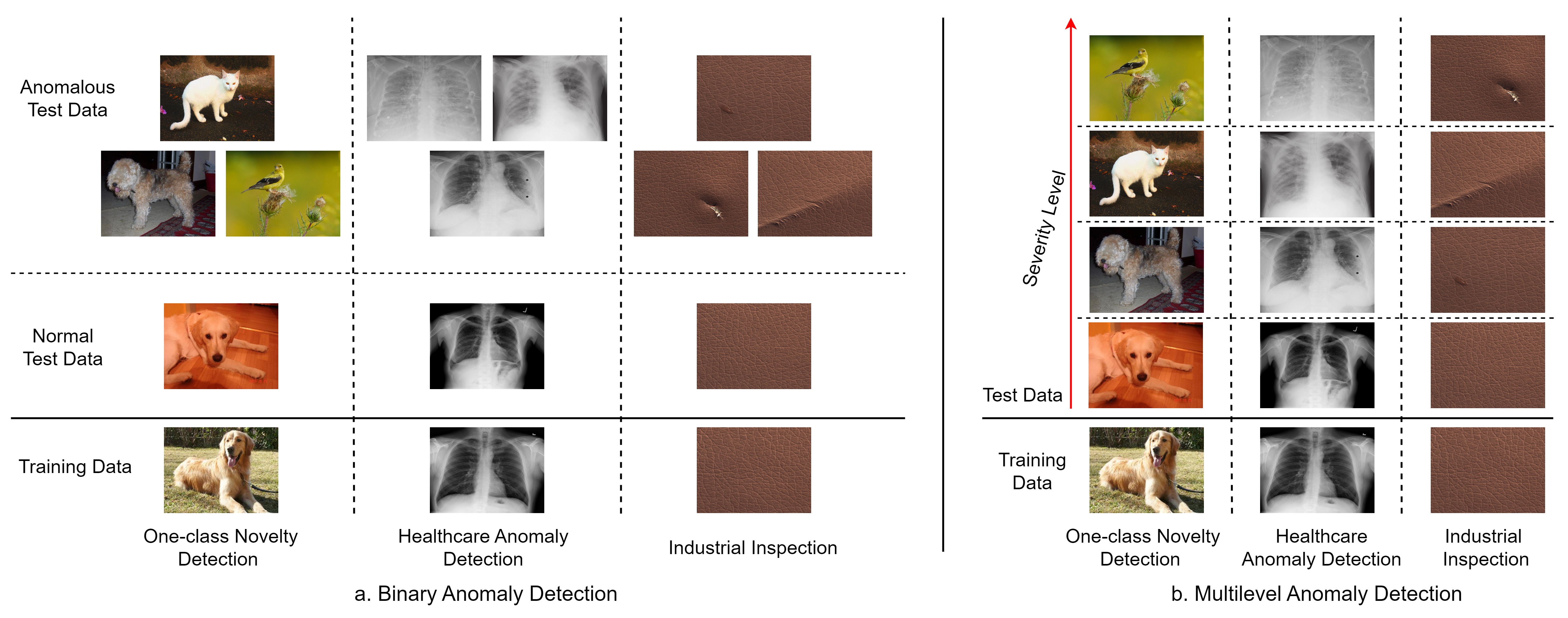}
    \caption{
(a) Binary Anomaly Detection classifies data as either in-distribution (ID) or out-of-distribution (OOD), without accounting for severity.
(b) The Multilevel Anomaly Detection setting categorizes OOD data by severity, reflecting the potential impact or risk. For instance, in COVID-19 chest X-rays, severity increases with greater lung involvement, from mild ground-glass opacities to extensive consolidation, indicating heightened clinical urgency.}
    \label{fig:intro}
\end{figure*}
Anomaly detection (AD) is a fundamental machine learning task that identifies deviations from normal patterns, with critical applications in domains such as one-class novelty detection \cite{ruff2018deep, chen2022deep}, industrial inspection \cite{bergmann2019mvtec, zou2022spot}, and medical diagnostics \cite{schlegl2019f, shvetsova2021anomaly}. However, in many real-world scenarios, anomalies are not uniformly significant; they exist on a spectrum of severity, ranging from minor anomalies that pose little risk to severe abnormalities that demand immediate attention. \textit{In a practical context, severity refers to the degree of potential impact or risk an anomaly poses to the system}. For instance, in industrial inspection, as shown in Figure \ref{fig:intro}b, anomalies range from minor surface contaminations to severe tears, requiring different levels of response. The ability to differentiate between these levels of severity is important for effective decision-making, prioritizing critical issues, and resource allocation. 

While distinguishing between different levels of anomaly severity is crucial, it remains unclear whether current anomaly detection methods are truly effective in capturing this distinction. Existing approaches typically focus on ensuring that the anomaly scores of anomalous data are higher than those of normal samples, operating primarily in a binary setting. Current techniques, such as reconstruction-based \cite{park2020learning,yan2021learning,zavrtanik2021reconstruction,pourreza2021g2,zaheer2020old}, one-class classifiers \cite{chen2022deep,goyal2020drocc,wu2019deep,sabokrou2020deep,yi2020patch}, and knowledge distillation-based models \cite{salehi2021multiresolution, wang2021student, deng2022anomaly, tien2023revisiting}, rely on metrics like reconstruction errors, distance measures, or likelihood estimates to flag deviations from normal patterns. However, the anomaly scores produced by these models are typically based on how far a data point deviates from normal data, but it is uncertain how well these scores correlate with the practical severity of the anomalies. 
% For instance, reconstruction-based and knowledge-distillation models rely heavily on reconstruction errors or prediction discrepancies, but are often overly sensitive to certain features or regions of the data, potentially misrepresenting the severity of the anomaly. 
% In some cases, minor deviations in critical areas lead to large anomaly scores, while more severe anomalies in less critical areas may only cause small changes. 

% Similarly, one-class classifiers like One-Class SVM \cite{scholkopf2001estimating} and Deep-VDD \cite{tax2004support} use distance from the decision boundary to score anomalies, but in high-dimensional spaces, this distance does not always correlate with severity due to issues like the curse of dimensionality or feature embedding artifacts.
% , and what practical applications could benefit from this assessment
These observations raise a key question: How accurately can the anomaly scores measure practical severity? To the best of our knowledge, there are no works that directly address this question. While \textit{near-distribution novelty detection} \cite{mirzaei2022fake} is related, it focuses on distinguishing between normal samples and abnormal samples that closely resemble normal ones rather than investigating the relationship between anomaly scores and severity. In this paper, we aim to study the relationship between anomaly score and severity of anomalies by presenting three main contributions:
\begin{itemize}
\item First, we propose a novel setting, \textbf{Multilevel Anomaly Detection (MAD)}, where the anomaly score should represent the severity of anomalies in real-world applications. We highlight the applications of multilevel anomaly detection in real-world scenarios, including one-class novelty detection, anomaly detection in medical imaging, and industrial inspection. 
% This also demonstrates the disconnect between anomaly scores from existing models and practical severity.
\item Building on the MAD setting, we introduce a new benchmark named \textbf{MAD-Bench}, designed to evaluate models on both their anomaly detection capabilities and their effectiveness in assigning severity-aligned anomaly scores. To achieve this, we adapt existing datasets from diverse domains to the multilevel anomaly detection context. In addition to conventional baselines, we incorporate Multimodal Large Language Model (MLLM)-based baselines, which leverage domain knowledge and reasoning abilities to assign anomaly scores. This makes MAD-Bench a comprehensive framework for evaluating model performance in MAD settings.
% , including one-class classification-based, knowledge distillation-based, reconstruction-based, memory-bank-based, synthetic-based, noisy-based and distribution map-based approaches
\item 
Finally, we conduct a comprehensive performance analysis using MAD-Bench, evaluating models on their ability to assign severity-aligned anomaly scores, the correlation between binary and multilevel detection performance, and the impact of anomalous feature area. We also assess performance across severity levels, the effect of including light anomalies in the normal class, and robustness under input corruption. These insights help improve AD models for practical severity alignment.
    % \item (Proposing method to improve the alignment between anomaly score and level of severity)
\end{itemize}

%% file: sec/2_related.tex
\section{Related Work}
\label{sec:formatting}

\textbf{Traditional Approaches}. Many approaches have been proposed for anomaly detection. One-class classification methods have progressed from machine learning approaches \cite{scholkopf2001estimating, tax2004support} to deep learning-based techniques \cite{ruff2018deep, chen2022deep, goyal2020drocc, wu2019deep, sabokrou2020deep, yi2020patch, liu2023simplenet}, which improve boundary refinement for distinguishing normal and anomalous data.  Reconstruction-based models \cite{yan2021learning, zavrtanik2021reconstruction, pourreza2021g2, zaheer2020old, zavrtanik2022dsr, liang2022omni, zavrtanik2021draem} identify anomalies by measuring deviations in reconstruction errors. Additionally, self-supervised learning techniques \cite{li2021cutpaste, yan2021learning, zhang2024realnet, schluter2022natural} introduce synthetic anomalies to enhance model differentiation, improving detection accuracy in complex scenarios.  Knowledge distillation-based methods employ a teacher-student framework \cite{bergmann2020uninformed, salehi2021multiresolution, wang2021student, deng2022anomaly, tien2023revisiting, gu2023remembering, zhang2023destseg}, where anomalies are detected based on deviations between teacher and student representations. Distribution map-based methods \cite{gudovskiy2022cflow, shin2023anomaly, Lei_2023_CVPR} model the normal data distribution and identify outliers as deviations. Memory-based approaches \cite{roth2022towards, bae2023pni, gu2023remembering, gong2019memorizing, hou2021divide, park2020learning, wang2021student} detect anomalies by comparing input features to representative normal data stored in a memory bank.

\textbf{MLLM-Based Approaches}. MLLMs have recently been leveraged for anomaly detection due to their ability to handle multimodal reasoning. These approaches often utilize zero-shot learning \cite{anovl, Jiang2024MMADTF} or few-shot learning \cite{li2024promptad, zhu2024toward}, allowing them to adapt to new datasets without extensive task-specific training.

\textbf{AD Benchmarks}. Several benchmarks for anomaly detection have been introduced \cite{Jiang2024MMADTF, ader, xie2024iad, akcay2022anomalib, han2022adbench, wang2024real, cao2023anomaly}. However, these benchmarks primarily focus on traditional binary anomaly detection rather than addressing more nuanced multilevel anomaly detection settings.

\textbf{Near out-of-distribution (OOD) detection} aims to identify anomalies that are similar to normal data,  showing only slight deviations \cite{mirzaei2022fake}. This differs from our work, which explores the relationship between anomaly scores and varying levels of severity.

\textbf{Multi-class AD} identifies anomalies across various object classes within a single, unified framework \cite{you2022unified, he2024diffusion, lu2023hierarchical, ader}. By training on multiple normal classes, it identifies anomalies as instances deviating from any learned normal patterns. In contrast, Multilevel AD focuses on representing multiple levels of anomalies during inference and ensures that anomaly scores reflect these levels.

%% file: sec/3_multiAD.tex
\section{Multilevel Anomaly Detection}
\subsection{Problem Formulation}
In anomaly detection, the goal is to define a function \( f: \mathcal{X} \to \mathbb{R} \) that assigns an anomaly score \( f(x) \) to each data point \( x \in \mathcal{X} \), where higher scores indicate a greater likelihood of being an anomaly. Traditional approaches typically treat anomalies in a binary manner, detecting whether a point is normal or anomalous. However, real-world anomalies often vary in severity rather than fitting into binary categories. For example, in medical diagnostics, doctors assess abnormalities on a spectrum, from mild issues needing observation to critical conditions requiring urgent care.

Motivated by this, we propose a \textit{Multilevel Anomaly Detection (MAD)} setting, where data points are from \( L_0 \) to \( L_n \), with \( L_0 \) representing the set of normal data and \( L_1, L_2, \dots, L_n \) representing sets of increasingly severe levels of anomaly.

Let the training set be defined as \( \mathcal{D}_{\text{train}} \subseteq L_0 \), consisting exclusively of normal data points from \( L_0 \). The testing set is defined as \( \mathcal{D}_{\text{test}} \subseteq L_0 \cup L_1 \cup \dots \cup L_n \), which includes data from all levels \( L_0, L_1, \dots, L_n \).

The goal is to design an anomaly scoring function \( f(x) \) such that higher severity levels correspond to higher anomaly scores. Specifically, normal data points in \( L_0 \) should be assigned the lowest anomaly scores \( f(x_0) \), while data points from higher severity levels \( L_1, L_2, \dots, L_n \) should receive correspondingly higher scores. This ensures a consistent and interpretable mapping between anomaly severity and anomaly scores.

% The anomaly scoring function \( f(x) \) is required to satisfy the property:

% \[
% f(x) \propto \text{level}(x),
% \]

% ensuring that for any two data points \( x_1, x_2 \in \mathcal{D}_{\text{test}} \), if \( \text{level}(x_1) > \text{level}(x_2) \), then:

% \[
% f(x_1) > f(x_2).
% \]

\subsection{Evaluation Protocol}

In this section, we outline the evaluation protocol used to assess the performance of models on the multi-level anomaly detection task. Three key metrics will be employed: \textbf{AUROC} \cite{hanley1982meaning}, \textbf{C-index} \cite{uno2011c}, and \textbf{Kendall’s Tau-b} \cite{kendall1938new}.

The AUROC metric is used to evaluate the model’s ability to distinguish between normal data and anomalies, which is commonly used in binary AD settings \cite{chen2022deep, cao2023anomaly}.

The C-index is a generalization of the AUROC that can evaluate how well anomaly scores align with the severity levels. In the context of MAD, C-index is defined as:
\[
C = \frac{\sum_{a=1}^{n} \sum_{b=0}^{a-1} \sum_{x_i \in L_a} \sum_{x_j \in L_b} \mathbbm{1}(f(x_i) > f(x_j))}{\sum_{a=1}^{n} \sum_{b=0}^{a-1} |L_a| \cdot |L_b|},
\]

Similar to the AUROC, a C-index of 1 corresponds to the best model prediction, while a C-index of 0.5 indicates a random prediction. A high C-index score indicates that the model correctly ranks higher-severity anomalies with higher anomaly scores.

To further assess the alignment between predicted anomaly scores and severity levels, we employ Kendall's Tau-b. Similar to C-index, Kendall's Tau-b evaluates the consistency between anomaly scores and severity levels but is stricter, as it requires samples within the same severity level to have identical anomaly scores to ensure perfect agreement. Kendall’s Tau ranges from \([-1, 1]\), with \(-1\) indicating a perfectly incorrect ranking, \(0\) signifying no correlation, and \(1\) representing a perfect severity ranking.
 The formula is detailed in the Appendix.

% This metric essentially checks whether the anomaly score assigned by the model is consistent with the true ordering of severity levels.
% This rank correlation coefficient measures the ordinal association between two variables—in this case, the model's anomaly score and the true severity level. Kendall's Tau is defined as:

% \[
% \tau = \frac{N_c - N_d}{\binom{n}{2}}
% \]

% where:
% \begin{itemize}
%     \item \(N_c\): the number of \textit{concordant pairs}, i.e., pairs \((x_i, x_j)\) where both anomaly scores and severity levels are ordered in the same direction.
%     \item \(N_d\): the number of \textit{discordant pairs}, i.e., pairs \((x_i, x_j)\) where anomaly scores and severity levels are ordered in opposite directions.
% \end{itemize}

% A higher Kendall’s Tau value implies a stronger correlation between the ranking of anomaly scores and the true severity levels.

% \vspace{-0.2cm}
\subsection{Applications of Multilevel Anomaly Detection}
Multilevel anomaly detection has significant implications across real-world domains where distinguishing anomalies based on severity is essential for effective decision-making. 

\textbf{One-class Novelty Detection} identifies samples that deviate from a defined normal class. For instance, a model trained on Golden Retriever (a dog breed) images can categorize anomalies by their deviation: Level 1 includes other dog breeds (minor deviation); Level 2, cats (moderate similarity); Level 3, birds (significant structural differences); and Level 4, flowers (inanimate objects). This framework enables one-class novelty detection to assign anomaly scores based on the degree of relevance to the trained class.

% \textbf{One-class Novelty Detection} focuses on identifying samples that deviate from a defined class of normal data. For instance, consider a model trained exclusively on images of Golden Retrievers (a dog breed). Using a multilevel anomaly detection framework, anomalies can be categorized based on their degree of deviation: Level 1 includes images of other dog breeds, representing minor deviations; Level 2 comprises images of cats, which share some structural similarities but are distinct from dogs; Level 3 consists of images of birds, which deviate significantly in structure and appearance; and Level 4 involves images of flowers, representing the highest level of deviation as unrelated inanimate objects. This framework enables one-class novelty detection to assign anomaly scores based on the degree of relevance to the trained class.

\textbf{Medical Imaging} often involves detecting and assessing anomalies that may significantly impact a patient’s health. Multilevel anomaly detection is valuable as it allows models to differentiate between conditions of varying severity. For instance, in a model trained on images of healthy skin, anomalies can be categorized into levels: Level 1 includes benign lesions, representing minor deviations with no immediate health risk; Level 2 encompasses precancerous lesions, indicating a moderate risk that requires monitoring or intervention; and Level 3 consists of cancerous lesions, severe anomalies demanding urgent medical attention. This structured detection enables healthcare professionals to prioritize high-risk cases, improving patient outcomes by addressing critical conditions promptly.

In \textbf{Industrial Inspection}, the economic and operational impact of anomalies can vary widely. Multilevel anomaly detection helps by assessing anomalies not just based on their physical characteristics, but also their potential economic or operational consequences. For example, a small cosmetic defect on a non-essential part may be a low-severity anomaly, while a malfunction in a critical machine component that leads to production downtime or safety hazards would be classified as a high-severity anomaly. This approach ensures that high-risk anomalies that pose significant economic or safety threats are addressed first, optimizing quality control and manufacturing processes.

By applying multilevel anomaly detection, real-world applications across these domains benefit from more precise anomaly classification, which enables better prioritization and resource allocation.

%% file: sec/4_benchmark.tex
\vspace{-0.2cm}
\section{MAD-Bench: Multilevel AD Benchmark}
% \vspace{-0.1cm}
\subsection{Datasets in MAD-Bench}
\vspace{-0.2cm}
\begin{table}[h]\footnotesize
\centering
\scalebox{0.90}{
\begin{tabular}{lcccc}
\toprule
\multirow{2}{*}{\textbf{Application}} & \multirow{2}{*}{\centering \textbf{Dataset}} & \multirow{2}{*}{\textbf{Subset}} & \multirow{2}{*}{\shortstack{\textbf{Total} \\ \textbf{Images}}} & \multirow{2}{*}{\shortstack{\textbf{Number} \\ \textbf{of Levels}}}  \\  
\\ \hline
\multirow{1}{*}{Novelty Detection} & \centering MultiDogs-MAD  & 5 & 9500 & 5   \\  
                             \hline
\multirow{2}{*}{Industrial Inspection} & \centering MVTec-MAD & 14 & 5073 & 4  \\  
                             & VisA-MAD  & 9 & 7874 & 4    \\\hline
\multirow{3}{*}{Medical Imaging} & \centering DRD-MAD & 1 & 4500 & 5   \\  
                             & Covid19-MAD  & 1 & 1364 & 7 \\
                             & SkinLesion-MAD  & 1 & 2469 & 4 
\\\bottomrule
\end{tabular}
}
\caption{Summary of the six datasets used in MAD-Bench, comprising a total of 31 subsets included in the benchmark.}
\label{tab:defect-datasets}
\end{table}
% , comprising a total of 31 subsets.
% \vspace{-0.3cm}
In our proposed benchmark, we evaluate anomaly detection models across various applications. Among the datasets used, two (DRD-MAD and Covid19-MAD) already include predefined severity labels. For the remaining datasets, which contain class labels for each sample. (e.g., defect types, disease types, or class names), we manually assign severity levels based on these class labels. Classes with ambiguity or samples that do not accurately reflect the assigned class name are excluded from the benchmark. Depending on the application, we categorize anomalies into different numbers of severity levels. The statistics of all datasets can be found in Table \ref{tab:defect-datasets}. Sample distribution across severity levels is detailed in the appendix.
% \vspace{-0.2cm}
\subsubsection{One-Class Novelty Detection Datasets}
% We introduce the \textbf{MultiDogs-MAD} dataset, designed to assess how well models detect anomalies based on class similarity. The training set consists of 500 samples from each of five dog species selected from Stanford Dogs Dataset \cite{dogsdataset}: Bichon Frise, Chinese Rural Dog, Golden Retriever, Labrador Retriever, and Teddy. Each dog species in Level 0 is sequentially designated as the normal class for training, with Levels 1 to 4 serving as testing datasets across all dog species. For further explanation, Level 0 represents normal samples from the selected dog breeds, which could be one of the breeds in Bichon Frise, Chinese Rural Dog, Golden Retriever, Labrador Retriever, and Teddy. Meanwhile, the testing phase uses datasets from Levels 1 to 4, each introducing progressively severe anomalies. Level 1 includes near-distribution anomalies from the other four dog breeds not chosen as the normal class in Level 0, Level 2 introduces moderate-severity anomalies with cat images (sourced from \citet{catdataset}), Level 3 incorporates higher-severity anomalies with bird images (sourced from \citet{birddataset}), and Level 4 consists of the highest-severity anomalies with flower images (sourced from \citet{flowerdataset}) that are entirely unrelated to animals. Each testing level comprises 500 samples.
We introduce the \textbf{MultiDogs-MAD} dataset to evaluate models' ability to detect anomalies based on class similarity. The training set consists of 500 samples from each of five dog breeds selected from the Stanford Dogs Dataset \cite{dogsdataset}: Bichon Frise, Chinese Rural Dog, Golden Retriever, Labrador Retriever, and Teddy. Each breed in Level 0 is sequentially used as the normal class, with levels 1 to 4 as testing datasets introducing anomalies of increasing severity. Level 0 represents normal samples from the selected breed. Level 1 includes near-distribution anomalies from other dog breeds not in the training set, Level 2 introduces cat images as moderate anomalies \cite{catdataset}, Level 3 includes bird images as high-severity anomalies \cite{birddataset}, and Level 4 contains flower images as the highest-severity anomalies \cite{flowerdataset}, unrelated to animals. Each testing level contains 500 samples.
% \vspace{-0.2cm}
\subsubsection{Industrial Inspection Datasets} 
Building on the MVTec \cite{bergmann2019mvtec} and VisA \cite{zou2022spot} datasets, we create two MAD datasets, MVTec-MAD and VisA-MAD, by assigning severity levels to anomalies based on the economic and operational impact of the class to which the anomalies belong. Classes that could not be confidently assigned a severity level were excluded. \textbf{MVTec-MAD} includes over 5,073 high-resolution images across fourteen object and texture categories, with defect-free images for training and diverse defects in the test sets. \textbf{VisA-MAD} contains 7,874 images in nine subsets, with 6,405 for training and 1,469 for testing. Both datasets include four severity levels: Level 0 (non-defect), Level 1 (minor defects that are easily repairable), Level 2 (moderate defects with some economic impact), and Level 3 (severe defects with high economic impact). 
% More details on the severity levels are provided in the Appendix.
% \vspace{-0.2cm}
\subsubsection{Medical Datasets} 
% We introduce three medical datasets, including DRD-MAD, Covid19-MAD, and SkinLesion-MAD.

The \textbf{DRD-MAD} is derived from the Diabetic Retinopathy Detection dataset \cite{diabetic-retinopathy-detection}, contains high-resolution retina images labeled on a 0-4 severity scale, where 0 represents no DR and 4 indicates proliferative DR. The images vary in quality and orientation due to different imaging conditions and camera models. For training, we randomly select 1000 normal samples (severity 0). Testing includes 700 samples for each severity level (0-4), ensuring balanced evaluation across all stages of the disease.

The \textbf{Covid19-MAD} is sourced from COVID-19 Severity Scoring \cite{danilov2022dataset}, contains 1,364 chest X-ray images: 580 COVID-19 positive cases and 784 normal cases, with 703 images allocated for training. Severity scores for positive cases range from 0 (no findings) to 6 (severe, above 85\% lung involvement), sourced from four public datasets and annotated by two independent radiologists.

The \textbf{SkinLesion-MAD} is designed to evaluate models in detecting and classifying skin anomalies across varying severity levels. It includes 500 training and 500 testing images of healthy skin sourced from Kaggle \cite{kaggle_skin}. Anomalous images from the ISIC Challenge 2018 dataset \cite{codella2019skin, tschandl2018ham10000}, labeled by two doctors through consensus, are divided into Benign lesions (e.g., Melanocytic nevus, Vascular lesions, 652 samples), Precancerous lesions (Actinic keratosis, 327 samples), and Cancerous lesions (Melanoma, 500 samples).

% \textbf{Skin-lesion (Skin-lesion-MLAD)} comprises both normal healthy images and anomalous images for training and testing purposes. For the normal healthy images, we utilized 500 images for training and another 500 images for testing, all sourced from Kaggle \cite{kaggle_skin}. These images were carefully inspected and then randomly cropped from larger images of healthy skin to create consistent datasets for both training and testing. For the anomalous data, we used images from the ISIC Challenge 2018 dataset \cite{codella2019skin, tschandl2018ham10000}, which were labeled by two doctors who reached a consensus on the classification. The dataset was subsequently divided into three categories: Benign lesions, which include Melanocytic nevus and Vascular lesions, with 652 samples; Precancerous lesions, specifically Actinic keratosis, with 327 samples; and Cancerous lesions, represented by Melanoma, with 500 samples.
% \vspace{-0.1cm}
\subsection{Conventional Baselines}

% \begin{table}[h]
% \centering
% \begin{tabular}{lll}
% \textbf{Method (Year)}          & \textbf{Type}      & \textbf{Datasets} \\ \hline
% IGD (2022) \cite{chen2022deep}  & OCC, Noi                       & (1)-(5)    \\   
% RD4AD (2022) \cite{deng2022anomaly}  & KD                         & (1)-(5)   \\ 
% PatchCore (2022) \cite{roth2022towards} & Mem                      & (2), (3) ?    \\ 
% CFLOW-AD (2022) \cite{gudovskiy2022cflow} & DM                      & (2), (3) ?     \\ \hline
% SimpleNet (2023) \cite{Liu_2023_CVPR} & OCC, Noi                   & (1)-(5)   \\ 
% RRD (2023) \cite{tien2023revisiting} & KD                         & (1)-(5)     \\ 
% OCR-GAN (2023) \cite{liang2023omni} & Reconstruction                         & (1)-(5)   \\ 
% PNI (2023) \cite{bae2023pni}   & Mem                            & (2), (3) ?   \\ 
% PyramidFlow (2023) \cite{lei2023pyramidflow} & DM                   & (2), (3) ?   \\ 
% SPR (2023) \cite{shin2023anomaly} & Other & (1)-(5)   \\\hline
% RealNet (2024) \cite{zhang2024realnet} & Reconstruction, Syn                  & (2), (3)   \\ 
% AE4AD (2024)\cite{cai2024rethinking} & Reconstruction                  & (4)-(6)
% \end{tabular}
% \caption{Summary of methods used in baselines, categorized by type of approach and applicable datasets.}
% \label{tab:datasets}
% \end{table}

% Deep SVDD \cite{ruff2018deep}, 
% synthetic-based (Syn), 
% SimpleNet \cite{Liu_2023_CVPR}
% RealNet \cite{zhang2024realnet},PyramidFlow \cite{lei2023pyramidflow}
We include baselines of several types, including one-class classification, knowledge distillation-based, reconstruction-based, memory-bank, and distribution map-based methods. The methods we evaluate are Skip-GAN \cite{akccay2019skip},  OCR-GAN \cite{liang2023omni}, AE4AD \cite{cai2024rethinking}, IGD \cite{chen2022deep}, RD4AD \cite{deng2022anomaly}, RRD \cite{tien2023revisiting},  PatchCore \cite{roth2022towards}, PNI \cite{bae2023pni}, CFLOW-AD \cite{gudovskiy2022cflow} and SPR \cite{shin2023anomaly}. The details are in Table \ref{tab:baselines}.

For all baselines, we strictly follow the original settings, including all hyperparameters such as the learning rate, number of training epochs, optimizer, and others. For new datasets, we apply baseline settings from similar datasets. 
For each model, we derive the sample-level anomaly score based on its design and use it to evaluate performance.

\begin{table}[h]\footnotesize
\centering
% \scalebox{0.87}{
\begin{tabular}{lll}
\toprule
\textbf{Type}      & \textbf{Method}          & \textbf{Year} \\ \hline
\multirow{3}{*}{Reconstruction} 
& Skip-GAN \cite{akccay2019skip}   & 2019   \\
& OCR-GAN \cite{liang2023omni}    & 2023   \\
& AE4AD \cite{cai2024rethinking}  & 2024   \\ \hline
\multirow{2}{*}{Knowledge Distillation} 
& RD4AD \cite{deng2022anomaly}    & 2022   \\
& RRD \cite{tien2023revisiting}   & 2023   \\ \hline
\multirow{2}{*}{Memory-bank} 
& PatchCore \cite{roth2022towards} & 2022   \\
& PNI \cite{bae2023pni}           & 2023   \\ \hline
\multirow{1}{*}{One-class Classification} 
& IGD \cite{chen2022deep}         & 2022   \\ \hline
\multirow{2}{*}{Distribution Map} 
& CFLOW-AD \cite{gudovskiy2022cflow} & 2022 \\
& SPR \cite{shin2023anomaly}        & 2023  \\ \hline
\multirow{4}{*}{MLLMs} 
& MMAD-4o                    & Ours   \\
& MMAD-4o-mini               & Ours   \\
& MMAD-Sonnet        & Ours   \\
& MMAD-Haiku           & Ours   \\
\bottomrule
\end{tabular}
% }
\caption{Summary of conventional baselines by approach type.}
\label{tab:baselines}
\end{table}
\vspace{-0.4cm}
% We conduct baselines across several approaches, including one-class classification (OCC), knowledge distillation-based (KD), Reconstruction-based (Rec), memory-bank (Mem), synthetic-based (Syn), noisy-based (Noi) and distribution map-based (DM) methods. The methods we evaluate are IGD \cite{chen2022deep}, SimpleNet \cite{Liu_2023_CVPR}, RD4AD \cite{deng2022anomaly}, RRD \cite{tien2023revisiting}, RealNet \cite{zhang2024realnet}, OCR-GAN \cite{liang2023omni}, PNI \cite{bae2023pni}, PatchCore \cite{roth2022towards}, PyramidFlow \cite{lei2023pyramidflow}, and CFLOW-AD \cite{gudovskiy2022cflow}. Note that some methods may belong to multiple categories, and not all of them can be applicable to all datasets as detailed in the Table \ref{tab:baselines}.
% \vspace{-0.1cm}
\subsection{MLLM-based Baselines}
% \vspace{-0.35cm}
\begin{figure}[!ht]
    \centering
    \includegraphics[width=0.49\textwidth]{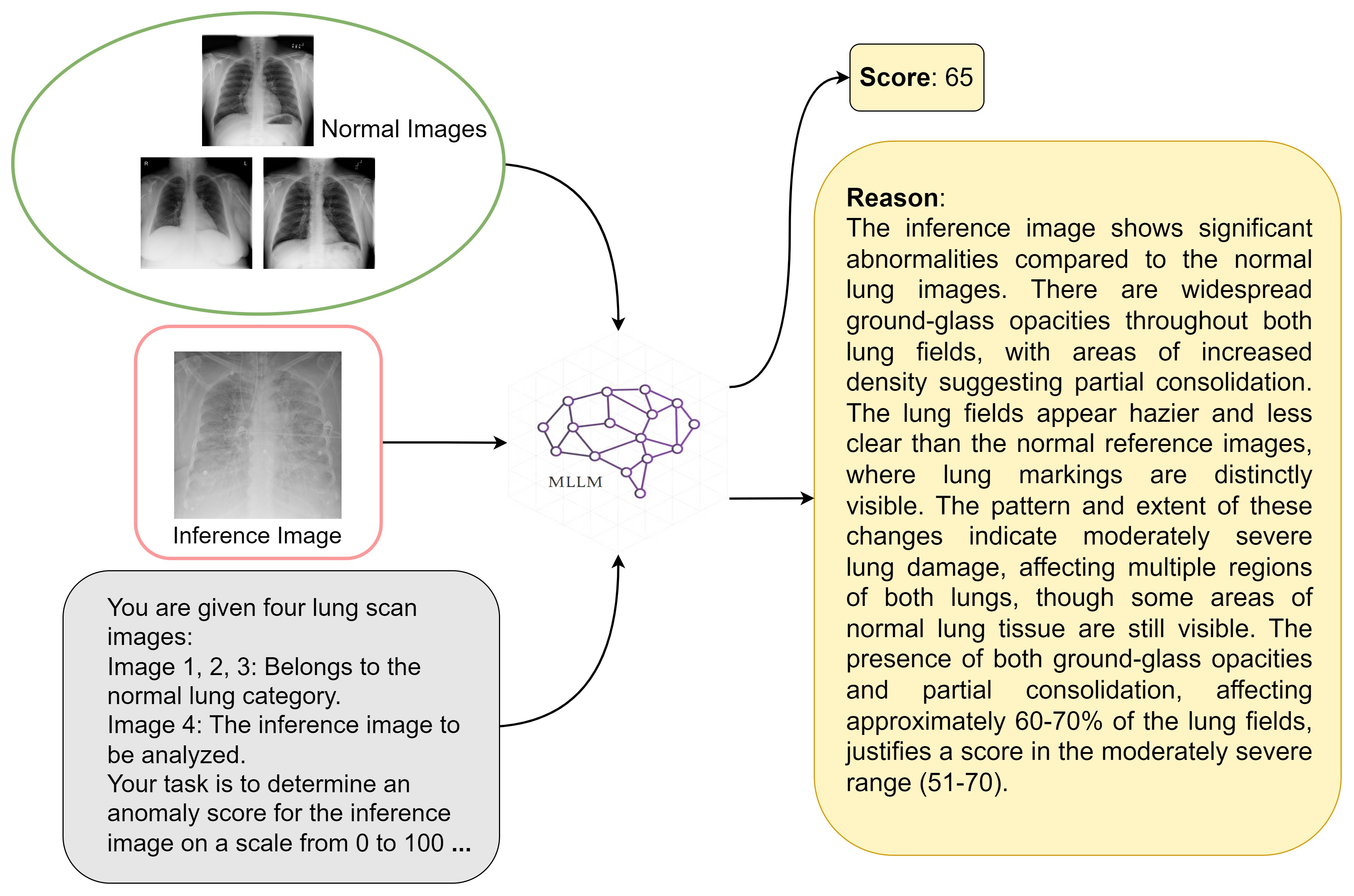}
% \vspace{-0.5cm}
    \caption{
MLLM-based baselines enable few-shot multilevel AD by leveraging their domain knowledge without fine-tuning, unlike conventional baselines requiring target data training.
    }
    \label{fig:MLLMs-based}
\end{figure}
% \vspace{-0.2cm}
We acknowledge that assigning anomaly scores corresponding to severity levels requires domain knowledge, which conventional baselines lack. Researchers have recently explored MLLMs for zero-shot learning \cite{anovl, Jiang2024MMADTF} or few-shot learning \cite{li2024promptad, zhu2024toward}, but these methods do not include the specific context of multilevel anomaly detection. Thus, we introduce MLLM-based baselines tailored to the MAD setting. We adopt a few-shot setting to balance effectiveness and cost, by utilizing four images: three normal images and one inference image. We prompt the model with the necessary application context, and instruct it to compare the inference image with the normal images. Based on this comparison, the model generates an anomaly score ranging from 0 to 100, accompanied by reasons, with higher scores indicating greater severity. Figure \ref{fig:MLLMs-based} illustrates the overall framework. We evaluate the approach using four state-of-the-art MLLMs: GPT-4o, GPT-4o-mini, Claude-3.5-Sonnet, and Claude-3.0-Haiku. Additional details on the experimental setup and prompts are provided in the appendix.

%% file: sec/5_expanalysis.tex
\section{Experiments and Analysis on MAD-Bench}
\subsection{Research Questions}
We conduct experiments to answer the research questions:
\begin{itemize}
\item \textbf{RQ1: (Benchmark and Model Types Analysis)} 
 How accurately can different types of anomaly detection models assign anomaly scores that align with severity levels across various applications?
\item \textbf{RQ2: (Binary-Multilevel Performance Correlation)} Does a model that performs well in binary anomaly detection also perform well in multilevel anomaly detection?
\item \textbf{RQ3: (Anomalous Area Effect)} How does the area of anomalies affect their anomaly score?
\item \textbf{RQ4: (Detection Performance Across Severity)} How does binary detection performance vary across different severity levels of anomalies ?
\item \textbf{RQ5: (Normal Class Expansion)} How do the models perform when light anomalies are considered acceptable and included as part of the normal class?
\item \textbf{RQ6: (Robustness Analysis}) How robust are detection models in aligning anomaly scores with severity under data corruption?
    % \item (\textbf{Practical guideline}) How can we adopt anomaly scores for measure severity in specific use cases ?
    % \item (\textbf{Future direction}) What is potential directions for improving the anomaly score-severity alignment ? 
\end{itemize}

\subsection{RQ1: Benchmark and Model Type Analysis}

\begin{table*}[htbp]\scriptsize
    \centering
    \renewcommand{\arraystretch}{1.2}
    \scalebox{0.9}{
    % \begin{tabular}{|p{2cm}p{0.4cm}<{\centering}p{0.4cm}<{\centering}p{0.4cm}<{\centering}
    %                 |p{0.4cm}<{\centering}p{0.4cm}<{\centering}p{0.4cm}<{\centering}
    %                 |p{0.4cm}<{\centering}p{0.4cm}<{\centering}p{0.4cm}<{\centering}
    %                 |p{0.4cm}<{\centering}p{0.4cm}<{\centering}p{0.4cm}<{\centering}
    %                 |p{0.4cm}<{\centering}p{0.4cm}<{\centering}p{0.4cm}<{\centering}
    %                 |p{0.4cm}<{\centering}p{0.4cm}<{\centering}p{0.5cm}<{\centering}
    %                 |p{0.3cm}<{\centering}p{0.3cm}<{\centering}p{0.3cm}<{\centering}}
\setlength{\tabcolsep}{2.85pt} % Adjust column separation
    \begin{tabular}{p{2cm}ccc|ccc|ccc|ccc|ccc|ccc|ccc|ccc}
    \toprule
        \multirow{2}{*}{\textbf{Method}} 
        & \multicolumn{3}{c}{\textbf{MultiDogs-MAD}} 
        & \multicolumn{3}{c}{\textbf{MVTec-MAD}}
        & \multicolumn{3}{c}{\textbf{VisA-MAD}}
        & \multicolumn{3}{c}{\textbf{DRD-MAD}}
        & \multicolumn{3}{c}{\textbf{Covid19-MAD}}
        & \multicolumn{3}{c}{\textbf{SkinLesion-MAD}}
        & \multicolumn{3}{c}{\textbf{Average}}
        & \multicolumn{3}{c}{\textbf{Average Rank}}\\
\cline{2-25}

        & \textbf{AUC} & \textbf{Ken} & \textbf{\ C} 
        & \textbf{AUC} & \textbf{Ken} & \textbf{\ C} 
        & \textbf{AUC} & \textbf{Ken} & \textbf{\ C} 
        & \textbf{AUC} & \textbf{Ken} & \textbf{\ C} 
        & \textbf{AUC} & \textbf{Ken} & \textbf{\ C} 
        & \textbf{AUC} & \textbf{Ken} & \textbf{\ C}
        & \textbf{AUC} & \textbf{Ken} & \textbf{\ C}
        & \textbf{AUC} & \textbf{Ken} & \textbf{\ C}\\
\cline{1-25}
    % Example row, replace with actual data
    Skip-GAN \cite{akccay2019skip} & 87.37 & \textbf{\textcolor{blue}{0.541}} & \textbf{\textcolor{blue}{80.25}} & 57.75 & 0.057 & 53.35 & 82.64 & 0.487 & 80.29 & 51.36 & 0.014 & 50.77 & 68.50 & 0.006 & 50.34 & 99.60 & 0.406 &	73.65 & 74.54 & 0.252 & 64.78 & 11 & 12 & 11\\
    OCR-GAN \cite{liang2023omni} & 75.63 & 0.343 & 69.19 & 94.39 & 0.401 & 73.90 & 91.17 & 0.526 & 83.03 & 53.83 & 0.054 & 53.03 & 65.46 & 0.038 & 52.09 & 99.53 & 0.391 & 72.76 & 80.00 & 0.292 & 67.33 & 10 & 11 & 9\\ %-6.38
    AE4AD \cite{cai2024rethinking} & 48.54 &  0.064 & 46.43 & 71.48 & 0.259 & 65.41 & 76.08 & 0.365 & 72.88 & 49.66 & 0.029 & 51.63 & 74.45 & 0.189 & 60.26 & 99.13 & 0.505 & 79.42 & 69.89 & 0.214 & 62.67 & 13 & 13 & 12\\ 
    RD4AD  \cite{deng2022anomaly}  & 67.44 & 0.308 & 67.20 & 98.92 & 0.504 & 79.86 & 95.24 & \textbf{\textcolor{blue}{0.640}} & \textbf{\textcolor{blue}{89.79}} & 61.30 & 0.217 & 62.14 & 83.48 & 0.219 & 61.90 & 98.94 & 0.509 & 79.60 & 84.22 & 0.399 & 73.41 & 8	& 7 & 6 \\
    RRD \cite{tien2023revisiting} & 81.67 & 0.510 & 78.51 & \textbf{\textcolor{blue}{99.51}} & \textbf{\textcolor{blue}{0.518}} & \textbf{\textcolor{blue}{80.72}} & 93.99 & 0.620 & 88.53 & 61.83 & 0.243 & 63.55 & 84.36 & 0.240 & 63.07 & 99.53 & \textbf{\textcolor{blue}{0.558}} & \textbf{\textcolor{blue}{82.48}} & 86.81	& \textbf{\textcolor{blue}{0.448}} & \textbf{\textcolor{blue}{76.14}} & 4 & \textbf{\textcolor{blue}{4}} & \textbf{\textcolor{blue}{3}}\\ 
    PatchCore \cite{roth2022towards} & 75.71 & 0.410 & 72.94 & 99.11 & 0.512 & 80.36 & 93.32 & 0.602 & 87.42 & 61.36 & 0.259 & 64.47 & 76.33 & 0.200 & 60.88 &  \textbf{\textcolor{blue}{100.0}} & 	0.456 & 76.57 & 84.31 & 0.407 & 73.77 &  6 &	6 & 5    \\ 
    PNI \cite{bae2023pni}  &  77.20 & 0.413 & 73.11 & 99.32 & 0.487 & 78.91 & \textbf{\textcolor{blue}{96.07}} & 0.622 & 88.64 & \textbf{\textcolor{blue}{62.50}} & \textbf{\textcolor{blue}{0.285}} & \textbf{\textcolor{blue}{65.94}} & \textbf{\textcolor{blue}{87.96}} & 0.241 & 63.12 & \textbf{\textcolor{blue}{100.0}} & 0.455 & 76.51 & \textbf{\textcolor{blue}{87.17}} & 0.417 & 74.37 & \textbf{\textcolor{blue}{2}} & 5 & 4\\ 
    IGD \cite{chen2022deep}& \textbf{\textcolor{blue}{95.02}} & 0.424 & 73.67 & 87.67 & 0.384 & 72.78 & 79.34 & 0.419 & 75.94 & 54.33 & 0.170 & 59.51 & 85.82 & \textbf{\textcolor{blue}{0.312}} & \textbf{\textcolor{blue}{66.99}} & 97.91 & 0.487 & 78.34 & 83.35 & 0.366 & 71.20 & 9 & 10 & 8 \\
    % SimpleNet \cite{Liu_2023_CVPR} & & & & & & & & & & & & & & & & & & \\ 
    CFLOW-AD \cite{gudovskiy2022cflow} & 89.45 & 0.431 & 74.11 & 96.79 & 0.466 & 77.58 & 85.08 & 0.493 & 80.43 & 60.21 & 0.238 & 63.30 & 74.03 & 0.180 & 59.78 &  99.98 &	0.389 & 72.63 & 84.26 & 0.366 & 71.30 & 7 & 9 & 7\\ 
    % PyFlow \cite{lei2023pyramidflow} & & & & & & & & & & & & & & & & & & \\ %-3.39
    SPR \cite{shin2023anomaly} & 64.29 & 0.242 & 63.55 & 97.76 & 0.451 & 76.78 & 56.05 & 0.087 & 55.49 & 46.32 &	0.003 & 48.11 & 55.14 & 0.006 & 53.51 & 91.31	& 0.335	& 69.50 & 68.48 & 0.191 & 61.16 & 14 & 14 & 13\\
    % RealNet \cite{zhang2024realnet} & & & & & & & & & & & & & & & & & & \\ 
    % NSA \cite{schluter2022natural} & & & & & & & & & & & & & & & & & & \\
    \hline
MMAD-4o & \textbf{\textcolor{red}{98.44}} & 0.933 & 95.91 & \textbf{\textcolor{red}{95.98}} & \textbf{\textcolor{red}{0.646}} & \textbf{\textcolor{red}{83.52}} & \textbf{\textcolor{red}{79.34}} & \textbf{\textcolor{red}{0.621}} & \textbf{\textcolor{red}{77.74}} & \textbf{\textcolor{red}{65.80}} & 0.433 & 67.72 & 88.07 & 0.547 & 76.94 & 99.43 & \textbf{\textcolor{red}{0.694}} & \textbf{\textcolor{red}{85.61}} & \textbf{\textcolor{red}{87.85}} & \textbf{\textcolor{red}{0.646}} & \textbf{\textcolor{red}{81.24}} & \textbf{\textcolor{red}{1}} & \textbf{\textcolor{red}{1}} & \textbf{\textcolor{red}{1}}\\ 
MMAD-4o-mini & 98.27 & 0.743 & 80.35 & 93.45 & 0.614 & 81.89 & 76.44 & 0.500 & 74.06 & 63.55 & 0.348 & 66.66 & 79.58 & 0.351 & 66.98 & 99.75 & 0.653 & 84.97 & 85.17 & 0.535 & 75.82 & 5 & 3 & 4\\
MMAD-Sonnet & 97.89 & \textbf{\textcolor{red}{0.936}} & \textbf{\textcolor{red}{97.34}} & 90.02 & 0.557 & 77.33 & 76.24 & 0.561 & 74.86 & 65.02 & \textbf{\textcolor{red}{0.403}} & \textbf{\textcolor{red}{69.30}} & \textbf{\textcolor{red}{92.35}} & \textbf{\textcolor{red}{0.601}} & \textbf{\textcolor{red}{80.54}} & 99.82 & 0.610 & 79.23 & 86.89 & 0.611 & 79.77 & 3 & 2 & 2\\
MMAD-Haiku & 93.03 & 0.767 & 88.55 & 76.10 & 0.366 & 66.63 & 61.81 & 0.326 & 60.43 & 53.39 & 0.125 & 56.20 & 59.42 & 0.142 & 54.16 & \textbf{\textcolor{red}{99.85}} & 0.479 & 74.19 & 73.93 & 0.367 & 66.69 & 12 & 8 &10\\

 % DDAD-ASR \cite{shin2023anomaly} & & & & & & & & & & & & & & & & & & \\ 
    % \hline
    \bottomrule
    
    \end{tabular}}
    \caption{Multilevel AD performance comparison across six datasets. The results are averaged across all subsets of each dataset. Higher AUROC (AUC) (\%), Kendall's Tau-b (Ken), and C-index (C) (\%) values indicate better performance. A lower average rank reflects better overall results based on the average scores for each metric. \textcolor{Red}{RED}: Best MLLM-based baseline. \textcolor{blue}{BLUE}: Best conventional baseline.  (RQ1-2)}
    \label{tab:benchmark}
\end{table*}

 We conducted a comprehensive benchmark across six datasets: MultiDogs-MAD, MVTec-MAD, VisA-MAD, DRD-MAD, Covid19-MAD, and SkinLesion-MAD, evaluating models on three metrics: AUROC (binary AD), C-index, and Kendall's Tau-b. Table \ref{tab:benchmark} shows results averaged across dataset subsets. 
 Detailed results are in the appendix.
 
 % The results are in Table \ref{tab:benchmark}, which are averaged across all subsets of each dataset. Detailed results are in appendix.

% Conventional models showed strong binary detection capabilities, particularly in datasets like MVTEC-MAD and SkinLesion-MAD, as reflected in high AUC values. However, their ability to measure severity levels is inconsistent, with lower Ken and C indices in datasets like Dogs-MAD, DRD-MAD, and VisA-MAD. Even in datasets with strong binary detection performance, such as SkinLesion-MAD, these models struggled to align anomaly scores with severity levels, highlighting a limitation in capturing nuanced severity levels.
Conventional models perform reasonably well in multilevel AD tasks on four datasets: MultiDogs-MAD, MVTec-MAD, Visa-MAD, and SkinLesion-MAD, with state-of-the-art methods achieving C scores above 80\%. However, these models struggle on the two medical datasets, DRD-MAD and Covid19-MAD, where C scores are around 65\%. This suggests that multilevel AD for medical images, such as X-rays and retinal fundus images, presents greater challenges for conventional models. Among conventional  approaches, knowledge distillation (e.g., \texttt{RRD}, \texttt{RD4AD}) and memory bank-based methods (e.g., \texttt{PatchCore}, \texttt{PNI}) demonstrated higher overall performance. Other approaches, particularly reconstruction-based models, show lower performance in both binary detection and severity level alignment.

Most MLLM-based models (except \texttt{MMAD-Haiku}) demonstrated consistently better performance than conventional methods across all datasets, except the VisA-MAD dataset.  They achieved both high AUC values and stronger alignment between anomaly scores and severity levels. For instance, in MultiDogs-MAD, MVTec-MAD and SkinLesion-MAD, these models not only matched conventional models in binary detection but also excelled in severity differentiation, as reflected in higher C and Ken values. A notable challenge is observed with the DRD-MAD dataset, which focuses on diabetic retinopathy detection. This dataset is particularly difficult for all models due to its requirement for expert-level knowledge to accurately assess severity levels. 

Generally, most MLLM-based models outperformed conventional models in multilevel anomaly detection by effectively leveraging domain knowledge, multimodal reasoning, and contextual information to better understand anomaly characteristics and severity.

% \begin{tcolorbox}[colframe=brown!100!black, colback=brown!10]
% \textbf{Finding 1}: MLLM-based models surpass conventional approaches in aligning anomaly scores with severity levels. Among conventional methods, knowledge distillation and memory bank-based approaches perform best.
% \end{tcolorbox}

\begin{tcolorbox}[colframe=brown!100!black, colback=brown!10]
\textbf{Finding 1}: Among conventional methods, knowledge distillation and memory bank-based methods better align anomaly scores with severity levels. However, MLLM-based models further outperform these methods, highlighting the importance of prior
knowledge in multilevel anomaly detection.
\end{tcolorbox}

\subsection{RQ2: Binary-Multilevel Performance Correlation}

Since most prior evaluations in anomaly detection focus on the binary setting, we investigate how model performance under the binary setting correlates with performance under our proposed multilevel setting. To this end, we compute the average scores for each metric across all datasets and rank the baselines accordingly, as shown in Table~\ref{tab:benchmark} (a lower average rank indicates better performance). We then calculate Spearman correlation coefficients \cite{zar2005spearman} of 0.973 between AUC and C, and 0.916 between AUC and Ken, indicating a strong positive correlation between binary and multilevel performance metrics.

However, despite this overall strong correlation, certain models exhibit notable discrepancies. For most models, the rank differences between AUC and Ken or AUC and C remain within 2, but \texttt{MMAD-Haiku} shows a rank difference of 4, and \texttt{PNI} has a difference of 3. Besides, the binary detection rankings of MLLM-based baselines are consistently higher (worse) than their multilevel detection rankings. This suggests that while MLLM-based models excel in aligning anomaly scores with severity levels (multilevel detection), their binary anomaly detection performance is comparatively weaker. These observations show that binary evaluation does not fully reflect multilevel performance, underscoring the need for our benchmark.

% These observations indicate that performance evaluated under the binary setting does not necessarily reflect performance in a multilevel setting, highlighting the necessity of our proposed benchmark.

% \begin{tcolorbox}[colframe=brown!100!black, colback=brown!10]
% \textbf{Finding 2}: Binary and multilevel AD performance are strongly correlated across all models.
% % Binary detection ranking of MLLM-based baselines is consistently worse than their multilevel detection ranking.
% \end{tcolorbox}

\begin{tcolorbox}[colframe=brown!100!black, colback=brown!10]
\textbf{Finding 2}: Binary and multilevel detection metrics generally correlate, but some models, notably MLLM-based baselines, perform better in multilevel evaluation. 
\end{tcolorbox}

\subsection{RQ3: Anomalous Area Effect}
\begin{table}[htbp]\scriptsize
\centering
\begin{tabular}{lcccc}
\toprule
\textbf{Model} & \multicolumn{2}{c}{\textbf{MVTec-MAD}} & \multicolumn{2}{c}{\textbf{ViSA-MAD}} \\
% \cline{2-5} \cline{6-9}
                & \textbf{Risk-based} &  \textbf{Area-based} &  \textbf{Risk-based}  & \textbf{Area-based} \\
\hline
OCR-GAN       & 73.90 & 85.85 & 83.03 & 86.76 \\
RD4AD          & 79.08 & 86.84 &  89.79 & 92.29 \\
RRD           & 80.72 & 87.18 & 88.53 & 91.41 \\
PatchCore      & 80.36 &  87.18 & 87.42 & 90.71 \\
PNI           & 78.91 & 88.75 & 88.64 & 92.90 \\
IGD           & 72.80 & 83.86 & 75.94 & 77.84 \\
CFLOW-AD       & 77.58  & 86.64  & 80.43 & 83.27 \\
MMAD-4o        & 83.52 & 89.05 & 77.74 & 78.17 \\
MMAD-4o-mini   & 81.89 & 84.60 & 74.06 & 74.24 \\
MMAD-Sonnet       & 77.33 & 82.73 & 74.86 & 75.11 \\
\hline
\end{tabular}
\caption{Risk-based and Area-based performance comparison is reported in C-index (\%). Conventional methods tend to be biased toward larger anomalous areas (RQ3).}
\label{tab:area-based}
\end{table}
\begin{figure*}[!ht]
    \centering
    \includegraphics[width=1\textwidth]{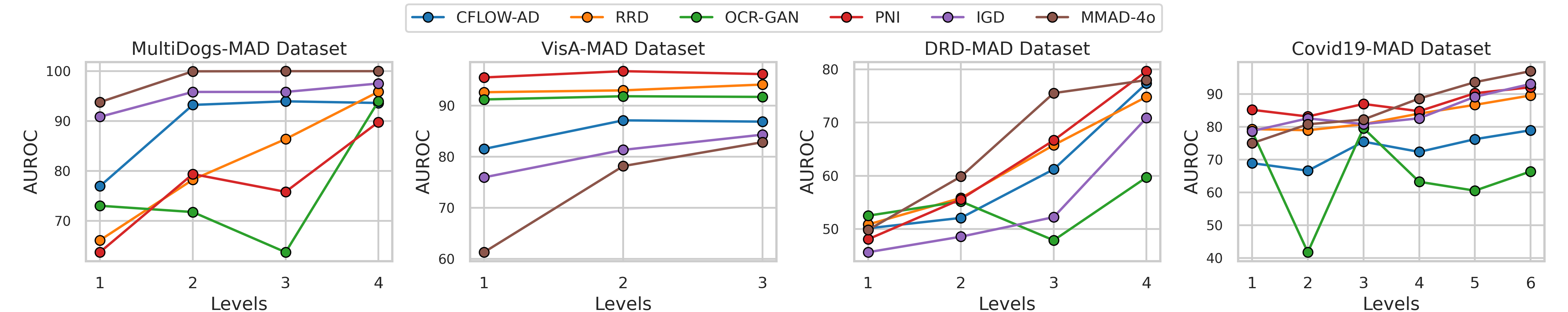}
    \caption{
Binary AD performance across severity levels. An upward trend is observed in most models (RQ4). 
    }
    \label{fig:increaseleve}
\end{figure*}
% \vspace{-0.2cm}
We hypothesize that the area of the anomalous region may affect the anomaly score generated by detection models, as these models often rely on differences in spatial features to identify anomalies. To validate this hypothesis, we conduct experiments using the MVTec-MAD and VisA-MAD datasets, as both include ground truth masks for the anomalies. The masks are resized to a width of 256 pixels while maintaining the aspect ratio. We then calculate the anomaly area for each sample and use it as the basis for defining an area-based severity level. 

Table \ref{tab:area-based} presents the multilevel anomaly detection performance for both risk-based severity levels (as used in our benchmark) and area-based severity levels. The experimental results indicate a strong correlation between the area of anomalous regions and anomaly scores. This is evidenced by the C values for area-based severity levels, which consistently exceed 80\% and are always higher than the C values for risk-based severity levels across both MVTec-MAD and VisA-MAD, with the exception of the MLLM-based baselines. Notably, models such as \texttt{RD4AD}, \texttt{PatchCore}, \texttt{RRD}, and \texttt{PNI} achieve exceptionally high C scores on VisA. These results suggest a bias in conventional models toward the area of anomalous regions. 

This bias can be advantageous in applications where severity levels are strongly correlated with the area of anomalous regions, such as surface scratches, where a larger area generally indicates greater severity. However, it can underestimate the severity of anomalies with small areas but significant impact, such as a severed electrical wire.

% \begin{tcolorbox}[colframe=brown!100!black, colback=brown!10]
% \textbf{Finding 3}: Conventional methods tend to be biased toward larger anomalous areas, by assigning them higher anomaly scores.
% \end{tcolorbox}
\begin{tcolorbox}[colframe=brown!100!black, colback=brown!10]
\textbf{Finding 3}: Conventional models are biased toward larger spatial anomalies, assigning them higher scores, while MLLMs show less of this tendency.
\end{tcolorbox}
\begin{figure*}[!ht]
    \centering
    \includegraphics[width=1\textwidth]{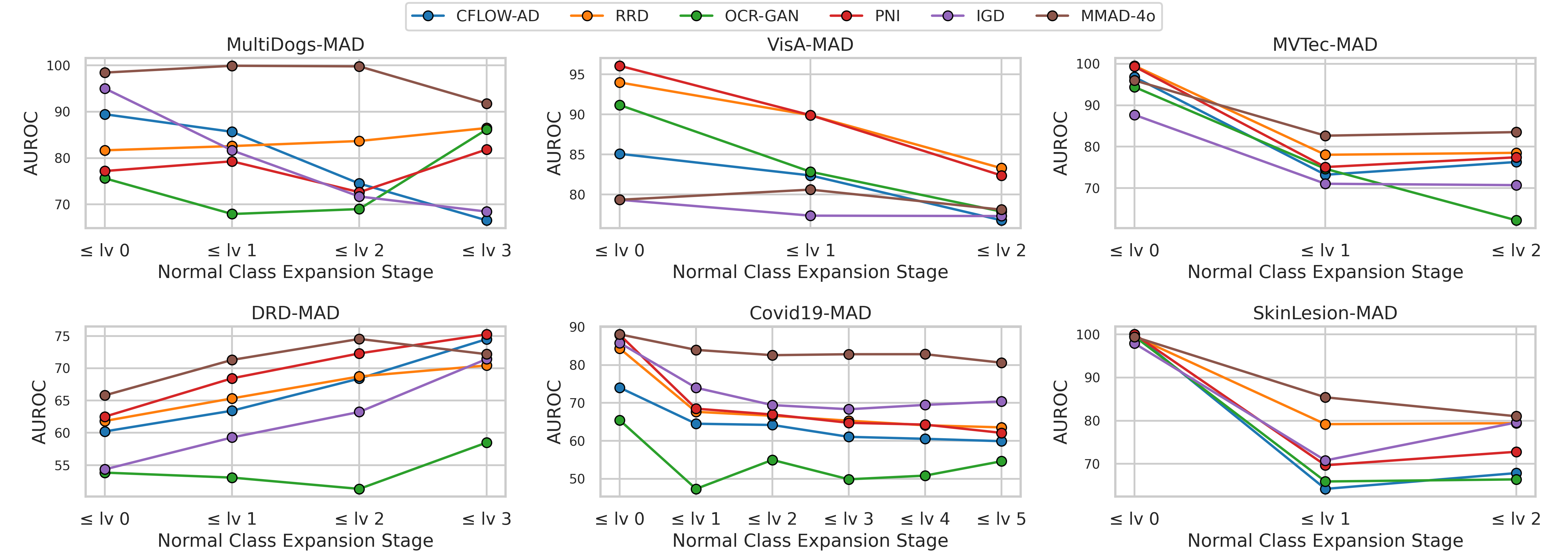}
    \caption{
Performance changes under expansions of the normal class definition. ``$\leq \text{lv $i$}$'' means that we consider the classes with severity levels no more than $i$ as normal classes in test time. Most models cannot maintain consistent performance across most datasets (RQ5). 
    }
    \label{fig:classexpand}
\end{figure*}
% \vspace{-0.2cm}
\subsection{RQ4: Detection Performance Across Severity}
A more severe anomaly is intuitively more noticeable to humans. We investigate whether models follow this intuition by examining their performance across increasing severity levels.
Specifically, we calculate the AUC between the normal class and each severity level. Results are reported only for MultiDogs-MAD, VisA-MAD, DRD-MAD, and Covid19-MAD in Figure \ref{fig:increaseleve}
, as performance on MVTec-MAD and SkinLesion-MAD is nearly flawless across all severity levels. Generally, most models exhibit an upward trend in detection performance as severity levels increase, indicating that higher-severity anomalies are detected more effectively. However, this trend is inconsistent at adjacent lower levels. Specifically, while the performance of most models at the highest levels across all datasets consistently surpasses that at lower levels, models such as \texttt{PNI} and \texttt{OCR-GAN} show inconsistencies in achieving better performance as the severity level increases at adjacent lower levels (e.g., Levels 1, 2, and 3).
% indicating potential instability in handling certain low-severity anomalies.
% \begin{tcolorbox}[colframe=brown!100!black, colback=brown!10]
% \textbf{Finding 4}: Generally, higher-severity anomalies are detected more effectively. However, this trend is not consistent for adjacent severity levels.
% \end{tcolorbox}
\begin{tcolorbox}[colframe=brown!100!black, colback=brown!10]
\textbf{Finding 4}: Most models detect higher-severity anomalies more effectively. Yet, fluctuations are observed at lower severity levels in certain cases.
\end{tcolorbox}
% Generally, the models show an upward trend in detection performance as the severity levels increase. This indicates that anomalies with higher severity levels are easier for the models to detect. For example, in the DRD-MAD and MultiDogs-MAD datasets, most models, such as MMAD-4o, RRD, and PNI, demonstrate consistently improved performance at higher severity levels. However, for less severe anomalies (e.g., Level 1 and Level 2), models like CFLOW-AD and OCR-GAN struggle, resulting in lower scores. Interestingly, the Covid19-MAD dataset shows a notable inconsistency for some models, such as OCR-GAN, which performs poorly at Level 2 before recovering at higher levels, indicating potential instability in handling certain low-severity anomalies. This trend suggests that while models are generally robust in detecting high-severity anomalies, their performance varies more significantly at lower severity levels.

\subsection{RQ5: Normal Class Expansion}
In practice, light anomalies are often considered acceptable and may be treated as part of the normal class. For example, a model trained on images of healthy skin without any abnormalities might encounter an inference image with a benign mole, which users want to consider as normal. To replicate this, during test time, we expand the definition of the normal class by progressively including samples from abnormal classes as normal based on severity levels. It is important to note that during the training phase, we exclusively use normal samples (level 0).

% The results in Figure \ref{fig:classexpand}
%  show that, overall, most models experience a decline in performance as the normal class expands to include light anomalies, except for MMAD-4o, which maintains stable performance on several datasets. This trend is observed across all datasets except DRD-MAD, where some models show more resilience or even slight improvements. The primary reason for this decline is that many models assign similarly high anomaly scores to light anomalies as they do to serious anomalies. This lack of differentiation between light and serious anomalies makes it challenging for the models to adjust to the expanded normal class, leading to a significant drop in anomaly detection performance. MMAD-4o’s robustness highlights its superior ability to distinguish between light and severe anomalies, enabling it to adapt better to these changes in normal class definitions. For DRD-MAD, the uptrend performance can be attributed to the nature of light anomalies, which are very challenging to distinguish from normal samples. As a result, models tend to assign low anomaly scores to these light anomalies, treating them as normal samples. This characteristic allows the models to benefit when light anomalies are included in the normal class, maintaining consistent detection performance despite the expanded normal definition.

Figure \ref{fig:classexpand}
 shows that all models experience a performance decline as the normal class expands to include light anomalies, except for \texttt{MMAD-4o}, which maintains stability on several datasets. This decline occurs because many models assign similarly high anomaly scores to light and serious anomalies, hindering adaptation to the expanded normal class. \texttt{MMAD-4o}'s robustness stems from its ability to distinguish between light and severe anomalies. For DRD-MAD, the upward trend is due to the nature of light anomalies, which are difficult to distinguish from normal samples. Models often assign low anomaly scores to these anomalies, treating them as normal. This behavior allows models to adapt more effectively when light anomalies are included in the normal class, maintaining detection performance.

\begin{tcolorbox}[colframe=brown!100!black, colback=brown!10]
\textbf{Finding 5}: 
All models experience a drop in performance when light anomalies are considered as normal samples, but MLLMs have a smaller drop.
\end{tcolorbox}

\subsection{RQ6: Robustness Analysis}

\begin{table}[htbp]\scriptsize
\centering
\setlength{\tabcolsep}{2.8pt} % Adjust column separation
\renewcommand{\arraystretch}{1} % Adjust row height
\begin{tabular}{lccc|ccc|ccc}
\toprule
\textbf{Model}& \multicolumn{3}{c}{\textbf{\ MultiDogs-MAD}} & \multicolumn{3}{c}{\textbf{\ MVTec-MAD}}  & \multicolumn{3}{c}{\textbf{SkinLesion-MAD}} \\
% \cline{2-5} \cline{6-10}

  & \textbf{Orig} & \textbf{Brgt} & \textbf{Noise} & \textbf{Orig} & \textbf{Brgt} & \textbf{Noise} & \textbf{Orig} & \textbf{Brgt} & \textbf{Noise}   \\
\hline
% RD4AD         & & & & & & & & &\\
OCR-GAN       & 69.19 & 66.70 & 68.08 & 73.90 & 69.33 & 67.23 & 72.76 & 70.02 & 63.38 \\
RRD           & 78.51 & 77.99 & 73.70 & 80.72 & 77.60 & 73.61 & 82.48 & 80.46 & 48.40 \\
PatchCore     & 72.94 & 73.40 & 69.41 & 80.36 & 78.91 & 78.66 & 76.57 & 75.86 & 70.75 \\
PNI           & 73.11 & 73.21 & 71.77 & 78.91 & 75.32 & 67.70 & 76.51 & 73.83 & 70.29 \\
IGD           & 73.67 & 68.70 & 71.59 & 72.78 & 65.91 & 65.91 & 78.34 & 80.44 & 67.55 \\
CFLOW-AD      & 74.11 & 72.44 & 69.41 & 77.58 & 67.02 & 63.48 & 72.63 & 70.43 & 68.13 \\
% CFLOW-AD      & 74.11 & 72.44 & 69.41 & 77.58 & 67.02 & 63.48 & 72.63 & 70.43 & 68.13 \\
MMAD-4o       & 95.91 & 94.28 & 94.71 & 83.52 & 79.00 & 72.82 & 85.61 & 80.91 & 54.72 \\
% MMAD-4o-mini  & & & & & & & & & \\
MMAD-Sonnet   & 97.34 & 96.19 & 95.04 & 77.33 & 73.64 & 67.16 & 79.23 & 77.33 & 67.43 \\
\hline
\end{tabular}
\caption{Performance comparison on brightness (Brgt), Gaussian noise (Noise), and origin (Orig) using C-index (\%). All models are adversely affected by corruption (RQ6).}
\label{tab:robustness}
\end{table}
% \vspace{-0.2cm}
To evaluate the robustness of anomaly detection models under corrupted input data, we use C-index on three datasets (MultiDogs-MAD, MVTec-MAD, and SkinLesion-MAD) under two types of input corruptions \cite{hendrycks2019robustness, cao2023anomaly}: brightness adjustment and noise addition. Table \ref{tab:robustness} shows that all models are negatively impacted by corruption, with noise corruption having a particularly strong effect on MVTec-MAD and SkinLesion-MAD. This is because noise can be misinterpreted by the models as defects, leading to a significant drop in C-index. In contrast, models evaluated on the MultiDogs-MAD are less affected by both types of corruption, due to the dataset's characteristic as a one-class novelty detection application, which encourages the models to rely on abstract features rather than fine-grained features.
\begin{tcolorbox}[colframe=brown!100!black, colback=brown!10]
\textbf{Finding 6}: 
All models are negatively affected by corruption, particularly on datasets requiring fine-grained feature analysis.
\end{tcolorbox}
% \vspace{-0.6cm}

%% file: sec/6_future.tex
\vspace{-0.2cm}
\section{Future Direction}
A promising direction for multilevel anomaly detection lies in integrating MLLMs with conventional approaches to leverage their complementary strengths. MLLMs excel in aligning anomaly scores with severity levels through domain knowledge and contextual reasoning, while conventional models perform better in binary detection, particularly for industrial inspection. A hybrid approach, such as a two-stage framework where conventional models detect anomalies and MLLMs assign severity scores, offers a robust and comprehensive solution by combining high accuracy in detection with contextual understanding for severity assessment.
% \vspace{-0.2cm}
% A promising direction for advancing multilevel AD is the integration of MLLM-based methods with conventional approaches. MLLMs excel in aligning anomaly scores with severity levels through domain knowledge and contextual reasoning, while conventional models often outperform in binary detection tasks, especially in the context of industrial inspection. By combining these strengths, a hybrid approach can address the limitations of each. For instance, conventional models can be leveraged for efficient feature extraction and binary anomaly detection, while MLLMs can handle severity-level reasoning and contextual anomaly assessment. This integration, such as a two-stage framework where conventional models detect anomalies and MLLMs assign severity scores, would  offer a robust and adaptable solution for multilevel AD.

%  with their strong feature extraction capabilities and lightweight inference

%% file: sec/7_conclusion.tex
\section{Conclusion}
% In conclusion, this paper introduces the novel Multilevel Anomaly Detection (MAD) setting, emphasizing severity-aligned anomaly scores for real-world applications. Through the development of MAD-Bench, we provide a robust benchmark to evaluate both anomaly detection and severity alignment capabilities. Our comprehensive analysis highlights the strengths and limitations of current models in the MAD context, offering valuable insights for advancing anomaly detection systems to better address practical severity considerations.

In this paper, we introduce Multilevel Anomaly Detection (MAD), a novel setting that emphasizes the alignment of anomaly scores with practical severity. To support this setting, we develop MAD-Bench, a comprehensive benchmark that evaluates the ability of models to reflect severity in diverse real-world applications. Through extensive analysis, we provide various findings, revealing insights into the relationship between binary and multilevel detection performance, as well as model robustness. These contributions highlight the potential for MAD to bridge the gap between theoretical anomaly detection and its practical deployment, paving the way for more effective and context-aware AD systems in real-world domains.

%% file: sec/X_apptmp.tex
% \clearpage

\setcounter{page}{1}

% \maketitlesupplementary
% \onecolumn

\section{Evaluation Metrics}
\subsection{AUROC metric}
Area Under the Receiver Operating Characteristic (AUROC) \cite{hanley1982meaning} in the context of Multilevel Anomaly Detection (MAD) is defined as:
\[
\text{AUROC} = \frac{\sum_{a=1}^{n} \sum_{x_i \in L_a} \sum_{x_j \in L_0} \mathbbm{1}(f(x_i) > f(x_j))}{\sum_{a=1}^{n} |L_a| \cdot |L_0|},
\]
\text{where:}
\begin{itemize}
    \item \( n \) is the number of severity levels.
    \item \( L_0 \) is the set of normal samples.
    \item \( L_a \) (for \( a = 1, 2, \ldots, n \)) are the sets of anomalous samples, grouped by severity level.
    \item \( f(x) \) is the anomaly score function.
    \item \( \mathbbm{1}(\cdot) \) is the indicator function, equal to 1 if the condition inside is true and 0 otherwise.
    \item \( |L_a| \) and \( |L_0| \) are the cardinalities (sizes) of the sets \( L_a \) and \( L_0 \), respectively.
\end{itemize}
AUROC achieves a perfect score of 1 if \textit{all the anomaly scores of abnormal samples are greater than all the anomaly scores of normal samples.}
 \subsection{C-index metric}
The C-index \cite{uno2011c} is a generalization of the AUROC that can evaluate how well anomaly scores align with the severity levels. In the context of MAD, C-index is defined as:
\[
C = \frac{\sum_{a=1}^{n} \sum_{b=0}^{a-1} \sum_{x_i \in L_a} \sum_{x_j \in L_b} \mathbbm{1}(f(x_i) > f(x_j))}{\sum_{a=1}^{n} \sum_{b=0}^{a-1} |L_a| \cdot |L_b|},
\]

A C-index of 1 corresponds to the best model prediction, achieved when \textit{all samples from higher-severity levels are consistently assigned higher anomaly scores than samples from lower-severity levels or normal samples}. A C-index of 0.5 indicates a random prediction. 

\subsection{Kendall’s Tau-b metric}
In our paper, in addition to the C-index, we employ Kendall's Tau-b \cite{kendall1938new} for evaluating the consistency between anomaly scores and severity levels. Kendall's Tau-b, a stricter version of Kendall's Tau, is specifically chosen because it accounts for tied ranks, which is critical when samples within the same severity level are expected to have identical anomaly scores if we expect a perfect consistency (i.e., $\tau_b = 1$).
% Specifically, it is designed to handle \textit{ties} (i.e., pairs where the anomaly scores or severity levels have the same values). 

Suppose a pair of samples \((x_i, x_j)\) is concordant if it follows the same order in terms of severity levels and anomaly scores. That is, if:
\begin{enumerate}
    \item The severity level of sample \(x_i\) is greater than that of sample \(x_j\), and the anomaly score of \(x_i\) is also greater than that of \(x_j\) or if
    \item The severity level of sample \(x_i\) is less than that of sample \(x_j\), and the anomaly score of \(x_i\) is also less than that of \(x_j\).
\end{enumerate}
The pair is discordant if it is in the reverse ordering for severity levels and anomaly score, or the values are arranged in opposite directions. That is, if:
\begin{enumerate}
    \item The severity level of sample \(x_i\) is greater than that of sample \(x_j\), but the anomaly score of \(x_i\) is less than that of \(x_j\) or if
    \item The severity level of sample \(x_i\) is less than that of sample \(x_j\), but the anomaly score of \(x_i\) is greater than that of \(x_j\).
\end{enumerate}
The two observations are tied if the severity level of sample \(x_i\) is equal that of \(x_j\) and/or the anomaly score of sample \(x_i\) is equal that of \(x_j\).

Kendall's Tau-b is formulated as:

\[
\tau_b = \frac{C - D}{\sqrt{(C + D + X_0)(C + D + Y_0)}},
\]

where:
\begin{itemize}
    \item \(C\): The number of concordant pairs,
    \item \(D\): The number of discordant pairs,
    \item \(X_0\): The number of pairs tied only on anomaly scores,
    \item \(Y_0\): The number of pairs tied only on severity levels.
\end{itemize}

Note that pairs where both the anomaly scores and severity levels are tied (\((XY)_0\)) are excluded from both the numerator and denominator of Kendall's Tau-b. This exclusion ensures that these pairs do not affect the metric's calculation or penalize the performance. 
% For example, if two samples have identical anomaly scores and belong to the same severity level, they contribute neither positively nor negatively to the rank correlation, as their ranking is already consistent. This property ensures that Kendall's Tau-b remains robust and interpretable, even in datasets with numerous ties.

The Kendall's Tau-b ensures that tied pairs only on anomaly scores or only on severity levels are not ignored but instead reduce the overall Kendall's Tau-b value to reflect the uncertainty caused by ties. Consequently, \(\tau_b\) will be equal to 1.0 when \textit{ the ordering of anomaly scores perfectly corresponds to the ordering of severity levels, and all samples within the same severity level have identical anomaly scores}.
 This characteristic makes Kendall's Tau-b particularly suitable in contexts where it is essential to ensure that a sample with a higher anomaly score always corresponds to a higher severity level. In contrast, the C-index can achieve a perfect score even when anomaly scores within the same severity level are not consistent, as it only considers pairwise comparisons across levels.

\section{Prompts and Design Choices for MLLM-based baselines}

\subsection{Setup}
For all experiments using MLLMs, we set the temperature to 0 to enable deterministic generation. The normal images are sourced from the training set and fixed for each subset.
\subsection{Prompt design}
We design prompts for MLLM-based baselines with the following key components, each carefully structured to guide the model's performance effectively:

\begin{itemize}
    \item \textbf{\textcolor{blue}{Context}}: Provides detailed information about the normal reference images and the inference images for comparison. This section ensures the model understands the baseline for normalcy and the target images to evaluate.
    
    \item \textbf{\textcolor{red}{Task Description}}: Clearly defines the objectives of the multilevel anomaly detection task. This includes outlining the model's role, such as identifying deviations between the reference and inference images and explaining the nature of potential anomalies.
    
    \item \textbf{\textcolor{purple}{Severity Levels Description}}: Describes the various severity levels of anomalies to guide the model in interpreting and giving corresponding anomaly scores. The prompt specifies the characteristics of each level to standardize interpretation.
    
    \item \textbf{\textcolor{brown}{Format Guidelines}}: Specifies the required structure and format for the model's response to ensure clarity and consistency. The output format includes the following components:
    \begin{itemize}
        \item Anomaly Score: A numerical value indicating the level of severity.
        \item Reasoning: A concise yet comprehensive explanation of the detected anomaly, providing insights into the specific features or conditions that led to the classification.
    \end{itemize}
\end{itemize}
Based on this design, five distinct prompts are provided, covering the following tasks: industrial inspection (MVTec-MAD and VisA-MAD), one-class novelty detection (MultiDogs-MAD), pulmonary imaging analysis (Covid19-MAD), diabetic retinopathy detection (DRD-MAD), and skin lesion detection (SkinLesion-MAD) (see text box below).

\section{Zero-shot AD and Few-shot AD using MLLMs}
\begin{table*}[htbp]\scriptsize
    \centering
    \renewcommand{\arraystretch}{1.2}
    \scalebox{1}{
\setlength{\tabcolsep}{2.85pt} % Adjust column separation

    \begin{tabular}{p{2cm}ccc|ccc|ccc|ccc|ccc|ccc|ccc}
    \toprule
        \multirow{2}{*}{\textbf{Method}} 
        & \multicolumn{3}{c}{\textbf{MultiDogs-MAD}} 
        & \multicolumn{3}{c}{\textbf{MVTec-MAD}}
        & \multicolumn{3}{c}{\textbf{VisA-MAD}}
        & \multicolumn{3}{c}{\textbf{DRD-MAD}}
        & \multicolumn{3}{c}{\textbf{Covid19-MAD}}
        & \multicolumn{3}{c}{\textbf{SkinLesion-MAD}}
        & \multicolumn{3}{c}{\textbf{Average}}\\
\cline{2-22}

        & \textbf{AUC} & \textbf{Ken} & \textbf{\ C} 
        & \textbf{AUC} & \textbf{Ken} & \textbf{\ C} 
        & \textbf{AUC} & \textbf{Ken} & \textbf{\ C} 
        & \textbf{AUC} & \textbf{Ken} & \textbf{\ C} 
        & \textbf{AUC} & \textbf{Ken} & \textbf{\ C} 
        & \textbf{AUC} & \textbf{Ken} & \textbf{\ C}
        & \textbf{AUC} & \textbf{Ken} & \textbf{\ C}\\
\cline{1-22}
    % Example row, replace with actual data
    % \hline
        RRD \cite{tien2023revisiting} & 81.67 & 0.510 & 78.51 & 99.51 & 0.518 & 80.72 & 93.99 & 0.620 & 88.53 & 61.83 & 0.243 & 63.55 & 84.36 & 0.240 & 63.07 & 99.53 & 0.558 & 82.48 & 86.81	& 0.448 & 76.14 \\ 
    PNI \cite{bae2023pni}  &  77.20 & 0.413 & 73.11 & 99.32 & 0.487 & 78.91 & 96.07 & 0.622 & 88.64 & 62.50 & 0.285 & 65.94 & 87.96 & 0.241 & 63.12 & 100.0 & 0.455 & 76.51 & 87.17 & 0.417 & 74.37 \\ 
\hline
    
    GPT-4o (zero-shot) & 98.67 & 0.942 & 98.44 & 82.16 & 0.525 & 75.11 & 68.70 & 0.422 & 68.22 & 50.25 & 0.044 & 50.29 & 50.95 & 0.064 & 50.63 & 61.33 & 0.180 & 55.77 & 68.68 & 0.363 & 66.41 \\
Sonnet (zero-shot) & 97.05 & 0.928 & 97.12 & 76.73 & 0.425 & 69.00 & 66.65 & 0.384 & 65.72 & 62.97 & 0.380 & 66.53 & 91.36 & 0.581 & 78.56 & 99.96 & 0.583 & 79.61 & 82.46 & 0.547 & 76.09 \\
\hline
MMAD-4o & 98.44 & 0.933 & 95.91 & 95.98 & 0.646 & 83.52 & 79.34 & 0.621 & 77.74 & 65.80 & 0.433 & 67.72 & 88.07 & 0.547 & 76.94 & 99.43 & 0.694 & 85.61 & 87.85 & 0.646 & 81.24 \\ 
MMAD-Sonnet & 97.89 & 0.936 & 97.34 & 90.02 & 0.557 & 77.33 & 76.24 & 0.561 & 74.86 & 65.02 & 0.403 & 69.30 & 92.35 & 0.601 & 80.54 & 99.82 & 0.610 & 79.23 & 86.89 & 0.611 & 79.77 \\
    \bottomrule
    
    \end{tabular}}
    \caption{Multilevel AD performance comparison between zero-shot and few-shot learning on MLLM-based baselines and state-of-the-art conventional baselines across six datasets. The results are averaged across all subsets of each dataset. Higher AUROC (AUC) (\%), Kendall's Tau (Ken), and C-index (C) (\%) values indicate better performance. Across most datasets, few-shot learning significantly outperforms zero-shot learning.}
    \label{tab:zeroshot}
\end{table*}

In addition to conducting experiments for few-shot settings on MLLMs (as reported in main paper), we also perform experiments under zero-shot settings. Specifically, in the zero-shot settings, we do not use normal images, aiming to evaluate the model's capability without reference images. The experimental results presented in Table \ref{tab:zeroshot}
 show that the performance of both Binary AD and Multilevel AD in zero-shot settings drops significantly compared to few-shot settings. This highlights the critical importance of using normal images as references in AD tasks.

\section{Output Examples of MLLM-Based Baselines}
This section provides examples of the output of MLLM-based baselines across datasets in Table \ref{table:example_mvtec}, \ref{table:example_multidogs} and \ref{table:example_covid}.

\section{Number of samples across Severity Levels}
We provide detailed number of samples across severity levels for each subset. The specifics are presented in Table \ref{tab:visa}, \ref{tab:mvtec}, \ref{tab:multidog}, and \ref{tab:medical}, corresponding to the details of VisA-MAD, MVTec-MAD, MultiDogs-MAD, and the three medical datasets.
\section{Full Results}
We provide detailed results for each subset across severity levels on six datasets. 
The detailed results for each dataset in the MAD-Bench framework are provided in the corresponding tables: VisA-MAD results are in Tables  \ref{tab:visa_detailed1} and \ref{tab:visa_detailed2}, MVTec-MAD results are in Tables \ref{tab:mvtec_detailed1}, \ref{tab:mvtec_detailed2}, \ref{tab:mvtec_detailed3}, MultiDogs-MAD results are in Table \ref{tab:multidogs_detailed},
 DRD-MAD results are in Table  \ref{tab:diabetic_detailed},
Covid19-MAD results are in Table  \ref{tab:covid19_detailed}
, and SkinLesion-MAD results are detailed in Table \ref{tab:skin-lesion_detailed}.

\begin{tcolorbox}[colback=gray!10, colframe=gray, width=1\textwidth,  boxrule=0.5mm, title= The prompt for Industrial Inspection (MVTec-MAD and VisA-MAD), fonttitle=\bfseries, coltitle=white, colbacktitle=gray]
\label{box:industrial}
\textbf{System prompt:}  \\You are a helpful assistant that responds in detecting anomalies on images in various context.\\

\textbf{Instruction:} \\
\textcolor{blue}{You are given four industrial inspection images:\\
Image 1, 2, 3: Belong to the non-defect category of \textcolor{blue}{\{subset\_name\}}.\\
Image 4: The inference image to be analyzed.}\\

\textcolor{red}{Your task is to determine an Anomaly Score for the inference image, reflecting the economic impact of the defect. A lower anomaly score suggests minimal economic impact, indicating that the item is likely usable or repairable, while a higher score indicates more severe impact, where the item may be unusable or costly to repair.}\\

\textcolor{purple}{Use the following guidelines and ranges to assign the anomaly score:\\
- If the inference image is consistent with the non-defect category, assign a score of 0.\\
- If the inference image shows minor defects that are unlikely to impact functionality and may be easily repairable, assign a score between 1 and 30.\\
- If the inference image has defects suggesting moderate economic impact, where the item is still usable but may require repair, assign a score between 31 and 70.\\
- If the inference image displays severe defects with high economic impact, suggesting the item may be unusable or expensive to repair, assign a score between 71 and 100.}\\

\textcolor{brown}{Output format:\\
Anomaly Score: [Score]\\
Reason: [Provide a brief explanation of why this score was assigned]\\
Please ensure the output strictly follows the format. You must answer the question even if you are not entirely certain about the answer.}

\end{tcolorbox}

\begin{tcolorbox}[colback=gray!10, colframe=gray, width=1\textwidth,  boxrule=0.5mm, title= The prompt for One-class Novelty Detection (MultiDogs-MAD), fonttitle=\bfseries, coltitle=white, colbacktitle=gray]
\label{box:multidogs}
\textbf{System prompt:}  \\You are a helpful assistant that responds in detecting anomalies on images in various context.\\

\textbf{Instruction:} \\
\textcolor{blue}{You are given four images:\\
Image 1, 2, 3: Belongs to the normal class \{\textcolor{blue}{subset\_name}\}.\\
Image 4: The inference image to be analyzed.}\\

\textcolor{red}{Your task is to determine an anomaly score for the inference image. This score should indicate how similar the inference image is to the class of the normal images. A lower anomaly score means the inference image is similar to the normal class, while a higher score indicates a greater deviation.
Take into account the relationship between the class of the normal images and the class of the inference image. Carefully consider that there may be different levels of relationships between the normal class and the various abnormal classes. For example, some abnormal classes might be more closely related to the normal class (e.g., two different breeds of cats), while others might be completely different (e.g., a cat versus a completely unrelated object). }\\

\textcolor{purple}{These relationships should reflect in the anomaly score:\\
- If the inference image belongs exactly to the same class as the normal images, assign a score of 0.\\
- If the inference image is similar but not identical to the normal class, assign a score between 1 and 50, with lower values indicating greater similarity.\\
- If the inference image belongs to an abnormal class but has some level of relationship to the normal class, assign a score between 51 and 75, considering the degree of similarity.\\
- If the inference image is completely unrelated to the normal class, assign a score between 76 and 100, with higher values indicating a greater dissimilarity.}\\

\textcolor{brown}{Output format: \\
Anomaly Score: [Score] \\
Reason: [Provide a brief explanation of why this score was assigned]\\
Please ensure the output strictly follows the format. You must answer the question even you are not sure about the anwser.}\\
\end{tcolorbox}

\begin{tcolorbox}[colback=gray!10, colframe=gray, width=1\textwidth,  boxrule=0.5mm, title= The prompt for DRD-MAD, fonttitle=\bfseries, coltitle=white, colbacktitle=gray]
\label{box:DRD}
\textbf{System prompt:}  \\You are a helpful assistant that responds in detecting anomalies on images in various context.\\

\textbf{Instruction:} \\
\textcolor{blue}{You are given four retinal images:\\
Image 1, 2, 3: Belongs to the normal retinal category (No DR). \\
Image 4: The inference image to be analyzed.}\\

\textcolor{red}{Your task is to determine an Anomaly Score for the inference image on a scale from 0 to 100, reflecting the extent of diabetic retinopathy (DR). A lower anomaly score suggests minimal or no DR, while a higher score indicates more severe stages of DR.}\\

\textcolor{purple}{Use the following guidelines to assign the Anomaly Score:\\
- If the inference image shows no signs of diabetic retinopathy, assign a score of 0.\\
- If the inference image shows mild signs of diabetic retinopathy, assign an anomaly score between 1 and 25.\\
- If the inference image has characteristics suggesting moderate diabetic retinopathy, assign an anomaly score between 26 and 50.\\
- If the inference image displays severe diabetic retinopathy, assign an anomaly score between 51 and 75.\\
- If the inference image reveals advanced proliferative diabetic retinopathy, assign an anomaly score between 76 and 100.}\\

\textcolor{brown}{Output format:\\
Anomaly Score: [Score]\\
Reason: [Provide a brief explanation of why this score was assigned]\\
Please ensure the output strictly follows the format. You must answer the question even if you are not entirely certain about the answer.} \\
\end{tcolorbox}

\begin{tcolorbox}[colback=gray!10, colframe=gray, width=1\textwidth,  boxrule=0.5mm, title= The prompt for Covid19-MAD, fonttitle=\bfseries, coltitle=white, colbacktitle=gray]
\label{box:Covid19}
\textbf{System prompt:}  \\You are a helpful assistant that responds in detecting anomalies on images in various context.\\

\textbf{Instruction:} \\
\textcolor{blue}{You are given four lung scan images:\\
Image 1, 2, 3: Belongs to the normal lung category. \\
Image 4: The inference image to be analyzed.}\\

\textcolor{red}{Your task is to determine an anomaly score for the inference image on a scale from 0 to 100, reflecting the extent of lung damage. A lower anomaly score suggests minimal or no lung damage, while a higher score indicates more severe lung damage.}\\

\textcolor{purple}{Use the following guidelines to assign the anomaly score:\\
- If the inference image is consistent with a normal lung, assign a score of 0.\\
- If the inference image shows very mild lung damage, such as isolated ground-glass opacities, assign a score between 1 and 15.\\
- If the inference image shows mild lung damage, with limited areas of ground-glass opacities, assign a score between 16 and 30.\\
- If the inference image has moderate lung damage, such as more widespread ground-glass opacities, assign a score between 31 and 50.\\
- If the inference image displays moderately severe lung damage, including partial lung consolidation, assign a score between 51 and 70.\\
- If the inference image shows severe lung damage, such as extensive consolidation in multiple lung regions, assign a score between 71 and 85.\\
- If the inference image reveals extreme lung damage, affecting more than 85 percent of the lungs, assign a score between 86 and 100.}\\

\textcolor{brown}{Output format:\\
Anomaly Score: [Score]\\
Reason: [Provide a brief explanation of why this score was assigned]\\
Please ensure the output strictly follows the format. You must answer the question even if you are not entirely certain about the answer.} \\
\end{tcolorbox}

\begin{tcolorbox}[colback=gray!10, colframe=gray, width=1\textwidth,  boxrule=0.5mm, title= The prompt for SkinLesion-MAD, fonttitle=\bfseries, coltitle=white, colbacktitle=gray]
\label{box:SkinLesion}
\textbf{System prompt:}  \\You are a helpful assistant that responds in detecting anomalies on images in various context.\\

\textbf{Instruction:} \\
\textcolor{blue}{You are given four dermoscopic images:\\
Image 1, 2, 3: Belong to the normal skin category.\\
Image 4: The inference image to be analyzed.}\\

\textcolor{red}{Your task is to determine an anomaly score for the inference image, reflecting its potential danger level in comparison to normal skin. A lower anomaly score suggests that the lesion is benign, while a higher score indicates a more dangerous lesion type, such as precancerous or cancerous.}\\

\textcolor{purple}{Use the following guidelines and ranges to assign the anomaly score:\\
- If the inference image is consistent with normal skin, assign a score of 0.\\
- If the inference image appears benign but shows minor atypical features, such as a mole (nevus) with slightly irregular borders, assign a score between 1 and 30 to indicate a benign lesion.\\
- If the inference image has characteristics suggesting it could be precancerous, assign a score between 31 and 70.\\
- If the inference image displays features strongly associated with malignancy, assign a score between 71 and 100, indicating a high risk for cancer.}\\

\textcolor{brown}{Output format:\\
Anomaly Score: [Score]\\
Reason: [Provide a brief explanation of why this score was assigned]\\
Please ensure the output strictly follows the format. You must answer the question even if you are not entirely certain about the answer. }\\
\end{tcolorbox}

\begin{table*}[h!]
\centering
\begin{tabular}{|m{6cm}|m{5cm}|m{5cm}|}
% \textbf{Image} & \textbf{Subset name} &\textbf{Severity Level (GT)} \\
\hline
Normal Image

\includegraphics[width=\linewidth]
{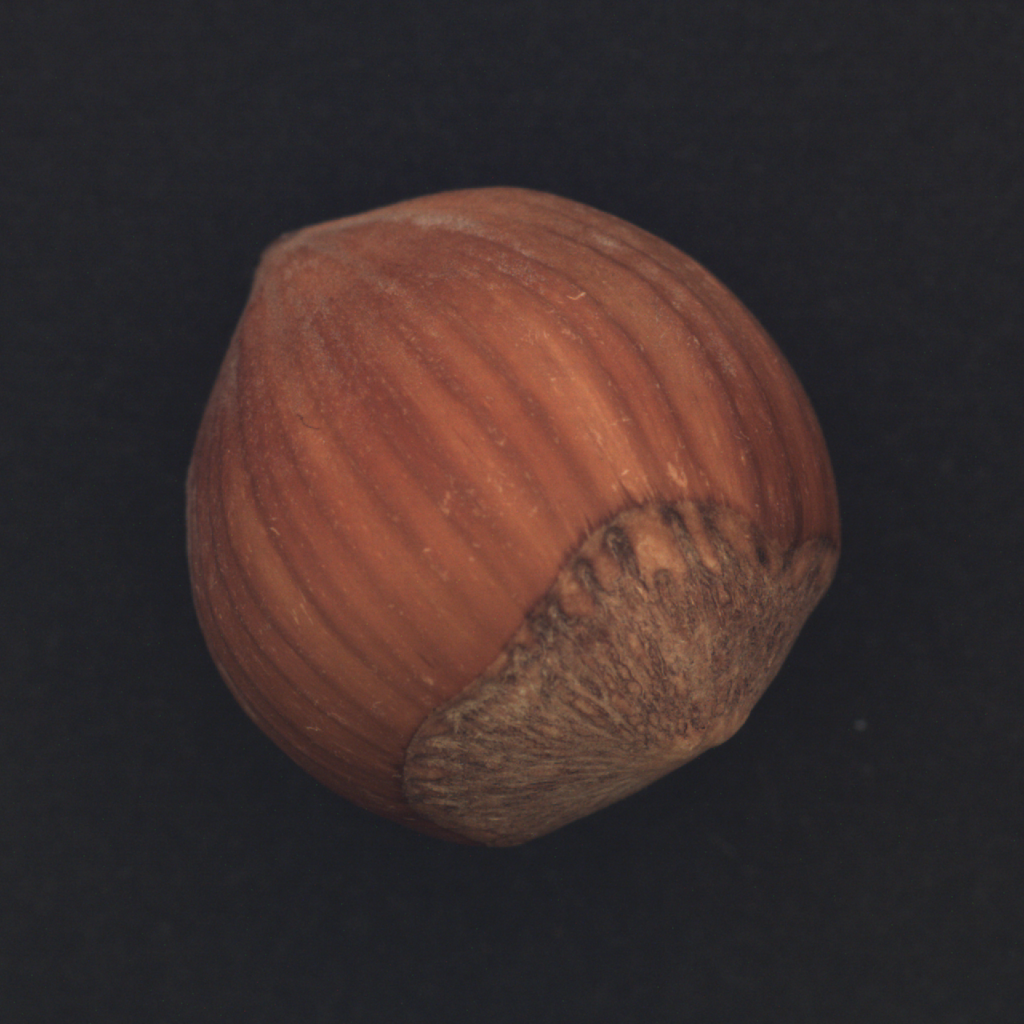} 

Inference Image

\includegraphics[width=\linewidth]
{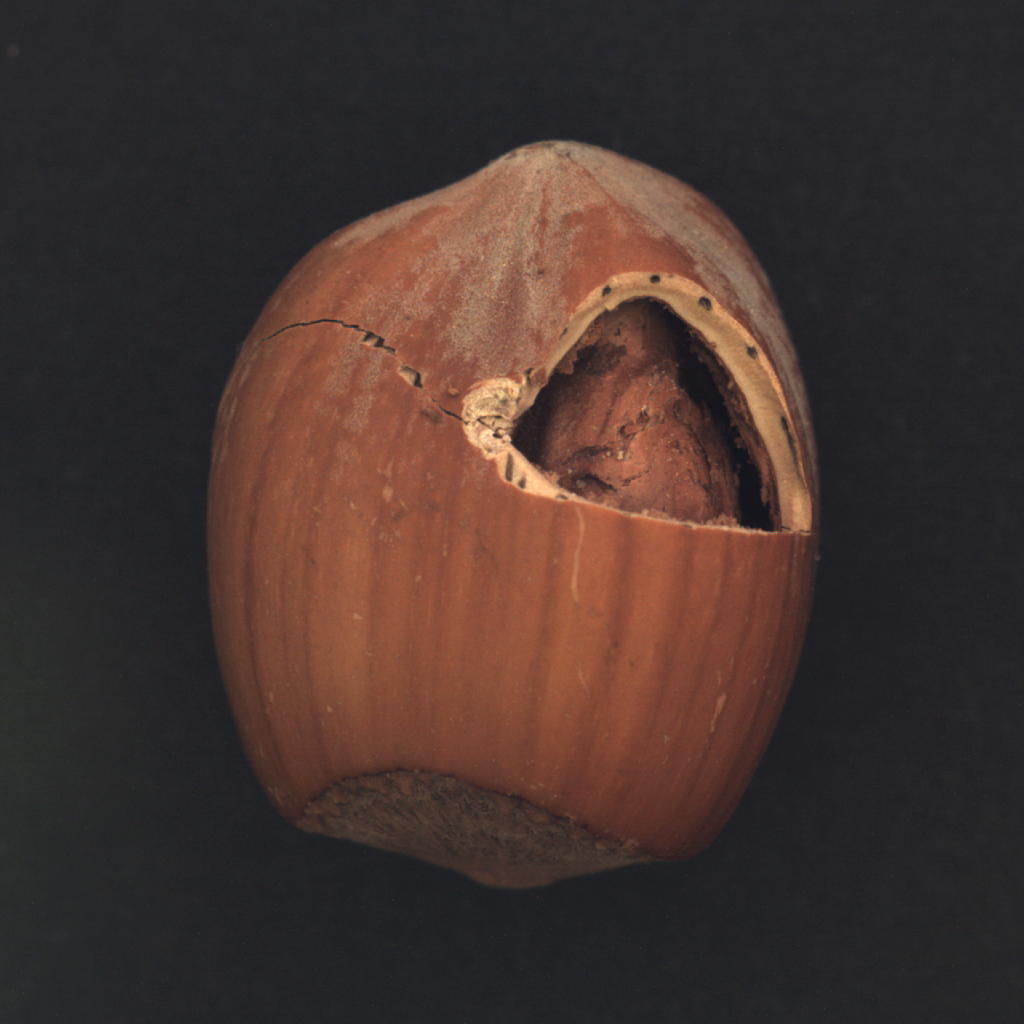} 
& \centering\arraybackslash \textbf{Subset:} Hazelnut in MVTec-MAD.  &  \centering\arraybackslash \textbf{Severity Level (GT): }3/3. The hazelnut has a large crack, indicating a high economic impact.
\\ \hline
Output from MMAD-4o  & \multicolumn{2}{m{10cm}|}{\textbf{Anomaly Score:} 71. 

\textbf{Reason:} The inference image shows a significant hole in the hazelnut, which is a severe defect. This type of damage likely renders the nut unusable or costly to repair, indicating a high economic impact.} \\
\hline
Output from MMAD-Sonnet  & \multicolumn{2}{m{10cm}|}{
\textbf{Anomaly Score:} 85. 

\textbf{Reason:}
The inference image shows a significant defect in the form of a hole or perforation in the hazelnut shell, along with what appears to be some discoloration around the damaged area. This type of damage likely indicates pest infestation or severe physical damage, which makes the hazelnut unsuitable for commercial sale or consumption. Such defects have a high economic impact as the product must be discarded due to food safety concerns and quality standards. Unlike minor surface blemishes, this type of structural damage cannot be repaired or salvaged, hence the high anomaly score.} \\
\hline
\end{tabular}
\caption{The output of MLLM-based baselines on a sample in MVTec-MAD dataset.}
\label{table:example_mvtec}
\end{table*}

\begin{table*}[h!]
\centering
\begin{tabular}{|m{6cm}|m{5cm}|m{5cm}|}
% \textbf{Image} & \textbf{Subset name} &\textbf{Severity Level (GT)} \\
\hline
Normal Image

\includegraphics[width=\linewidth]
{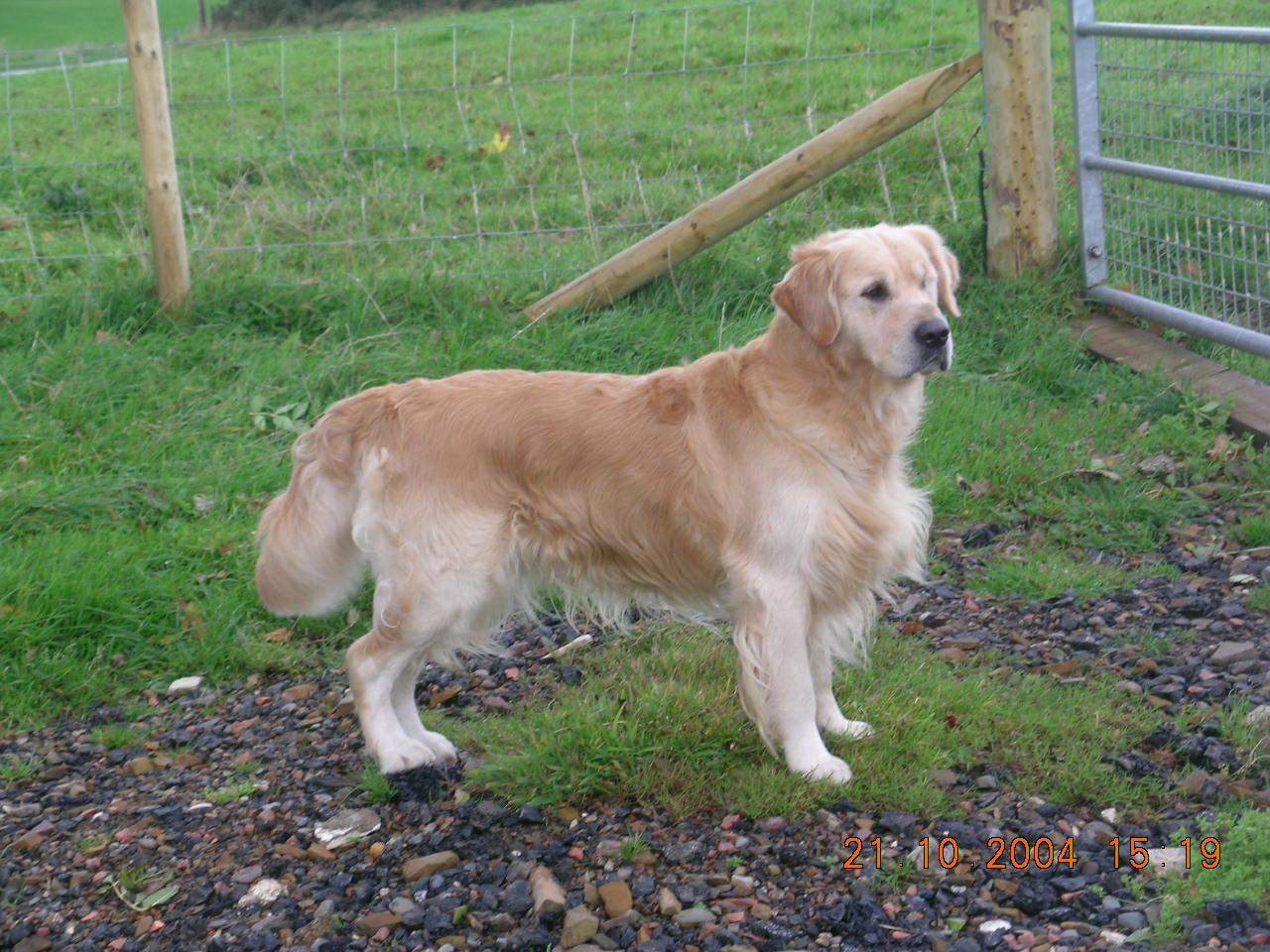} 

Inference Image

\includegraphics[width=\linewidth]
{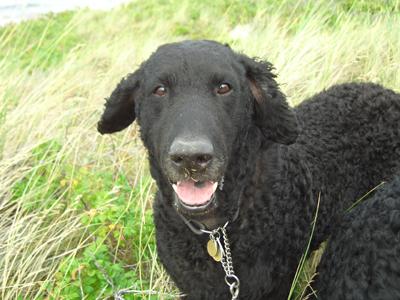} 
& \centering\arraybackslash \textbf{Subset Name:} Golden Retriever in MultiDogs-MAD (Normal images). The inference image depicts a dog of a different breed than the Golden Retriever.   &  \centering\arraybackslash \textbf{Severity Level (GT): }1/4. The dog is of a different breed than the Golden Retriever but is still a dog.
\\ \hline
Output from MMAD-4o  & \multicolumn{2}{m{10cm}|}{\textbf{Anomaly Score:} 51. 

\textbf{Reason:} The inference image depicts a dog that is not a golden retriever but is still a dog, indicating some level of relationship to the normal class. This results in a moderate anomaly score reflecting the similarity.} \\
\hline
Output from MMAD-Sonnet  & \multicolumn{2}{m{10cm}|}{
\textbf{Anomaly Score:} 35. 

\textbf{Reason:}
The inference image shows a Curly-Coated Retriever, which is different from but related to Golden Retrievers shown in the normal images. While it's not the same breed, it belongs to the same retriever family of dogs, sharing similar characteristics like body structure, general size, and purpose (retrieving dogs). The main differences are the black color and curly coat texture, as opposed to the golden straight coat of Golden Retrievers. The score of 35 reflects that while this is a different breed, it's still closely related to the normal class, being from the same retriever family, showing similar facial features, body structure, and behavioral characteristics typical of retriever breeds.} \\
\hline
\end{tabular}
\caption{The output of MLLM-based baselines on a sample in MultiDogs-MAD dataset.}
\label{table:example_multidogs}
\end{table*}

\begin{table*}[h!]
\centering
\begin{tabular}{|m{6cm}|m{5cm}|m{5cm}|}
% \textbf{Image} & \textbf{Subset name} &\textbf{Severity Level (GT)} \\
\hline
Normal Image 

\includegraphics[width=\linewidth]
{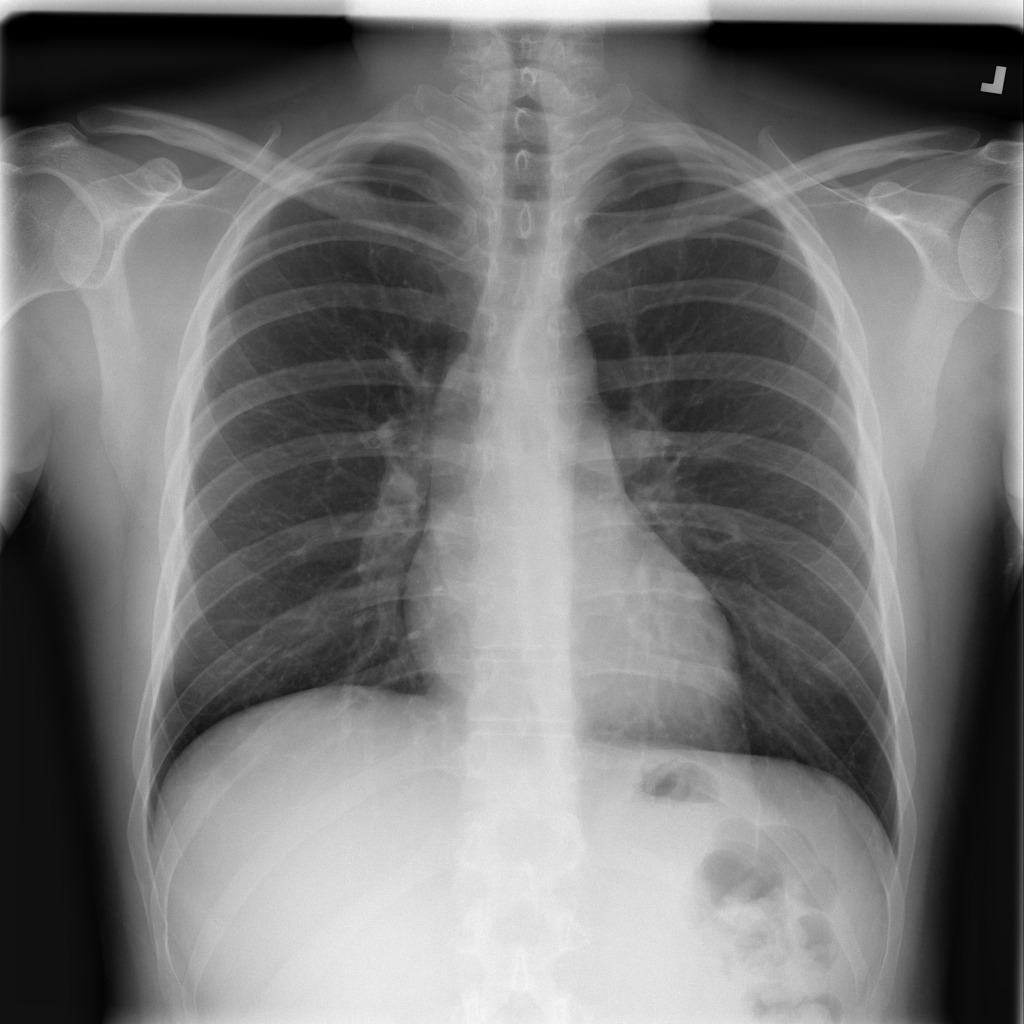} 

Inference Image

\includegraphics[width=\linewidth]
{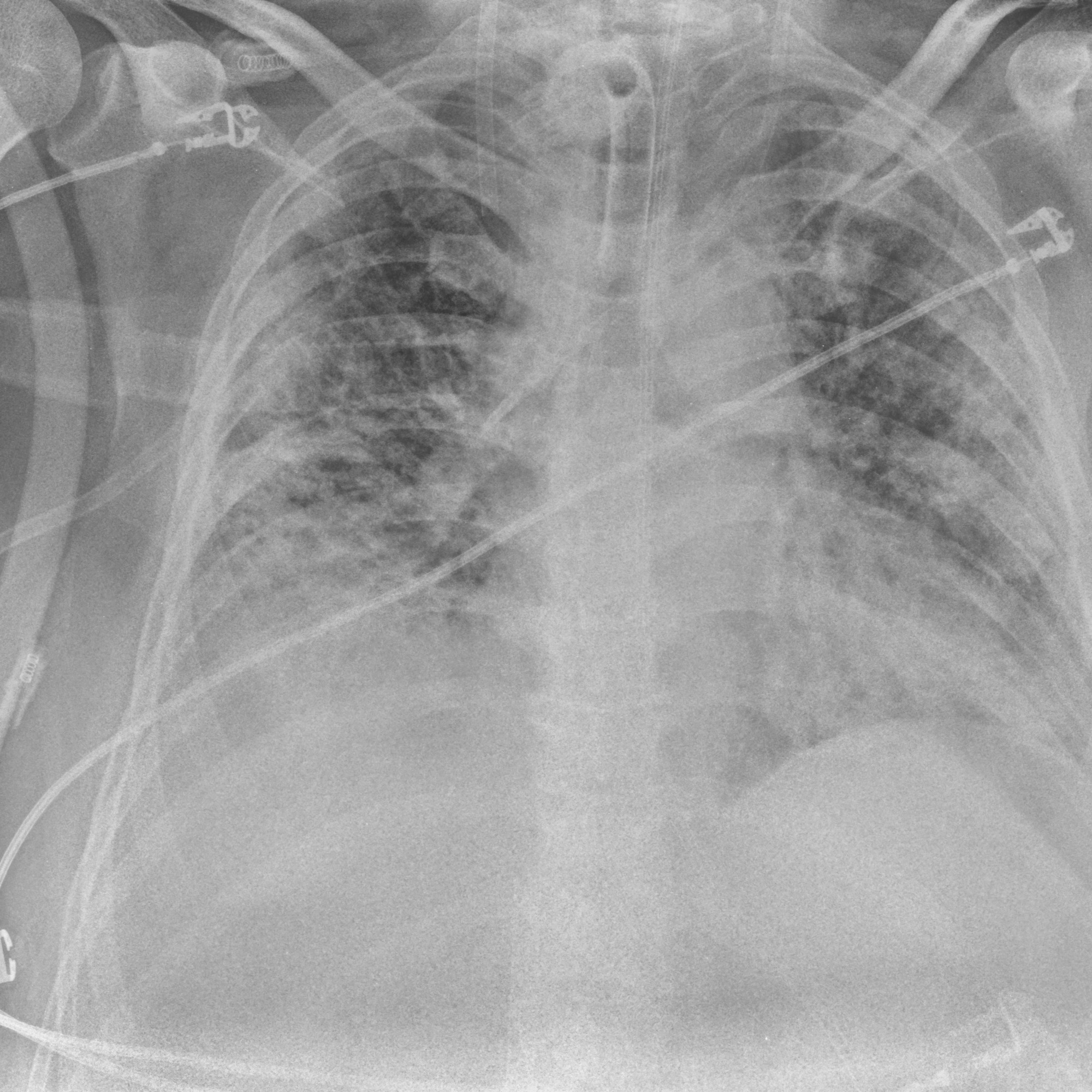} 
& \centering\arraybackslash \textbf{Dataset Name:} Covid19-MAD   &  \centering\arraybackslash \textbf{Severity Level (GT): }6/6. The inference image shows extreme lung damage, affecting more than 85\% of the lung.
\\ \hline
Output from MMAD-4o  & \multicolumn{2}{m{10cm}|}{\textbf{Anomaly Score:} 75. 

\textbf{Reason:} The inference image shows extensive consolidation in multiple lung regions, indicating severe lung damage. This is more pronounced compared to the normal lung images provided.} \\
\hline
Output from MMAD-Sonnet  & \multicolumn{2}{m{10cm}|}{
\textbf{Anomaly Score:} 75. 

\textbf{Reason:}
The inference image shows significant abnormalities compared to the normal lung images. There are extensive bilateral ground-glass opacities and consolidation patterns throughout both lung fields, particularly in the mid and lower zones. The lung fields appear much hazier and whiter compared to the clear, dark lung fields seen in the normal images. The presence of medical equipment (tubes\/lines) also suggests a severe clinical condition. The extent and distribution of the abnormalities affect approximately 75-80\% of the visible lung fields, placing this case in the severe category (71-85 range) of lung damage. While severe, it's not in the extreme category as there are still some areas of relatively preserved lung tissue visible.} \\
\hline
\end{tabular}
\caption{The output of MLLM-based baselines on a sample in Covid19-MAD dataset.}
\label{table:example_covid}
\end{table*}
\twocolumn
% \end{strip}
% \lipsum[3-4]
\begin{table*}[htbp]
    % \footnotesize
    \centering
    \renewcommand{\arraystretch}{1.2} % Adjust row height
    \scalebox{1}{ % Scale the table
    \setlength{\tabcolsep}{2.85pt} % Adjust column separation
    \begin{tabular}{lccccccccc}
        \toprule
        Level & Capsules & Chewinggum & Fryum & Macaroni1 & Macaroni2 & PCB1 & PCB2 & PCB3 & Pipe\_Fryum  \\
        \midrule
Level 0 (training) & 542 & 453 & 450 & 900 & 900 & 904 & 901 & 905 & 450 \\
Level 0 (test) & 60 & 50 & 50 & 100 & 100 & 100 & 100 & 101 & 50 \\
Level 1 & 45 & 40 & 6 & 16 & 16 & 14 & 14 & 20 & 39 \\
Level 2 & 20 & 8 & 35 & 12 & 9 & 66 & 67 & 60 & 15\\
Level 3 & 20 & 23 & 59 & 35 & 35 & 20 & 19 & 20 & 25\\\midrule
Total & 687 & 574 & 600 & 1063 & 1060 & 1104 & 1101 & 1106 & 579 \\
\bottomrule
    \end{tabular}}
    \caption{Number of samples of the VisA-MAD dataset across levels.}
    \label{tab:visa}
\end{table*}

\begin{table*}[htbp]
    \footnotesize
    \centering
    \renewcommand{\arraystretch}{1.2} % Adjust row height
    \scalebox{1}{ % Scale the table
    \setlength{\tabcolsep}{2.85pt} % Adjust column separation
    \begin{tabular}{lccccccccccccccc}
        \toprule
Level & Carpet & Grid & Leather & Tile & Wood & Bottle & Cable & Capsule & Hazelnut & Metal\_nut & Pill & Screw & Transistor & Zipper \\\midrule
Level 0 (training) & 280 & 264 & 245 & 230 & 247 & 209 & 224 & 219 & 391 & 220 & 267 & 320 & 213 & 240 \\
Level 0 (test) & 28 & 21 & 32 & 33 & 19 & 20 & 58 & 23 & 40 & 22 & 26 & 41 & 60 & 32 \\
Level 1 & 17 & 33 & 38 & 36 & 18 & 21 & 23 & 45 & 17 & 22 & 43 & 25 & 10 & 33 \\
Level 2 & 19 & 12 & 17 & 31 & 31 & 22 & 22 & 20 & 17 & 23 & 26 & 24 & 10 & 70\\
Level 3 & 34 & 12 & 37 & 17 & 11 & 20 & 11 & 44 & 36 & 25 & 17 & 24 & 20 & 16 \\\midrule
Total & 378 & 342 & 369 & 347 & 326 & 292 & 338 & 351 & 501 & 312 & 379 & 434 & 313 & 391 \\
\bottomrule
    \end{tabular}}
    \caption{Number of samples of MVTec-MAD dataset across levels.}
    \label{tab:mvtec}
\end{table*}

\begin{table*}[htbp]
    \footnotesize
    \centering
    \renewcommand{\arraystretch}{1.2} % Adjust row height
    \scalebox{1}{ % Scale the table
    \setlength{\tabcolsep}{2.85pt} % Adjust column separation
    \begin{tabular}{lccccc}
        \toprule
Level & Bichon Frise & Chinese Rural Dog & Golden Retriever & Labrador Retriever & Teddy \\\midrule
Level 0 (training) & 500 & 500 & 500 & 500 & 500 \\
Level 0 (test) & 500 & 500 & 500 & 500 & 500 \\
Level 1 & 500 & 500 & 500 & 500 & 500  \\
Level 2 & 500 & 500 & 500 & 500 & 500 \\
Level 3 & 500 & 500 & 500 & 500 & 500 \\
Level 4 & 500 & 500 & 500 & 500 & 500  \\\midrule
Total & 2500 & 2500 & 2500 & 2500 & 2500 \\
\bottomrule
    \end{tabular}}
    \caption{Number of samples of MultiDogs-MAD datasets across levels.}
    \label{tab:multidog}
\end{table*}

\begin{table*}[htbp]
    % \footnotesize
    \centering
    \renewcommand{\arraystretch}{1.2} % Adjust row height
    \scalebox{1}{ % Scale the table
    \setlength{\tabcolsep}{2.85pt} % Adjust column separation
    \begin{tabular}{lccc}
        \toprule
Level & Covid-MAD & DRD-MAD & SkinLesion-MAD \\\midrule
Leve 0 (training) & 703 & 1000 & 500\\
Level 0 (test) & 81 & 700 & 500\\
Level 1 & 96 & 700 & 642\\
Level 2 & 60 & 700 & 327\\
Level 3 & 62 & 700 & 500\\
Level 4 & 90 & 700 & None\\
Level 5 & 135 & None & None\\
Level 6 & 137 & None & None\\
\midrule
Total & 1364 & 4500 & 2469\\
\bottomrule
    \end{tabular}}
    \caption{Number of samples of three medical datasets across levels.}
    \label{tab:medical}
\end{table*}

\input{sec/tables/visa_detailed}

\input{sec/tables/mvtec_detailed}

\input{sec/tables/multidogs_detailed}

\input{sec/tables/diabetic_detailed}

\input{sec/tables/covid19_detailed}

\input{sec/tables/skin-lesion_detailed}

\twocolumn

%% file: sec/tables/visa_detailed.tex
\begin{table*}[htbp]\scriptsize
    \centering
    \renewcommand{\arraystretch}{1.2}
    \scalebox{1}{
\setlength{\tabcolsep}{2.85pt} % Adjust column separation
    \begin{tabular}{p{2cm}cccccc|cccccc}
    \toprule
        \multirow{3}{*}{\textbf{Method}} 
        & \multicolumn{4}{c}{\textbf{Binary AD performance}} 
        & \multicolumn{2}{c|}{\textbf{MAD performance}}
        & \multicolumn{4}{c}{\textbf{Binary AD performance}}
        & \multicolumn{2}{c}{\textbf{MAD performance}}\\
\cline{2-13}

        & \textbf{Level 1} & \textbf{Level 2} & \textbf{ Level 3} 
        & \textbf{Whole} & \textbf{\;Ken} & \textbf{C} 
        & \textbf{Level 1} & \textbf{Level 2} & \textbf{ Level 3} 
        & \textbf{Whole} & \textbf{\;Ken} & \textbf{C} \\
\cline{2-13}

        & \multicolumn{6}{c|}{\textbf{capsules}} 
        & \multicolumn{6}{c}{\textbf{chewinggum}}\\
\cline{1-13}
    % Example row, replace with actual data
    % \hline
        Skip-GAN \cite{akccay2019skip} & 63.11 & 79.58 & 71.08 & 68.86 & 0.286 & 67.12 & 82.55 & 92.00 & 96.00 & 87.97 & 0.560 & 83.84  \\
RD4AD  \cite{deng2022anomaly} & 85.26 & 98.33 & 98.50 & 91.45 & 0.701 & 91.93 & 95.30 & 100.0 & 100.0 & 97.35 & 0.736 & 94.45  \\
PatchCore \cite{roth2022towards} & 67.26 & 85.42 & 94.75 & 78.00 & 0.479 & 78.66 & 96.65 & 100.0 & 100.0 & 98.11 & 0.727 & 93.93  \\
CFLOW-AD \cite{gudovskiy2022cflow} & 61.41 & 94.00 & 87.58 & 75.24 & 0.416 & 74.89 & 98.00 & 100.0 & 100.0 & 98.87 & 0.714 & 93.14  \\
RRD \cite{tien2023revisiting} & 86.37 & 99.58 & 99.50 & 92.57 & 0.688 & 91.12 & 97.50 & 100.0 & 100.0 & 98.59 & 0.745 & 94.99  \\
OCR-GAN \cite{liang2023omni} & 89.48 & 84.50 & 80.08 & 86.10 & 0.310 & 68.51 & 91.80 & 92.75 & 94.96 & 92.93 & 0.572 & 84.56  \\
PNI \cite{bae2023pni} & 82.19 & 94.92 & 98.58 & 89.04 & 0.617 & 86.88 & 98.00 & 100.0 & 100.0 & 98.87 & 0.728 & 93.97  \\
SPR \cite{shin2023anomaly} & 67.19 & 85.08 & 74.42 & 73.10 & 0.311 & 68.60 & 35.95 & 31.75 & 30.26 & 33.63 & 0.213 & 37.13  \\
IGD \cite{chen2022deep} & 56.22 & 66.08 & 68.17 & 61.35 & 0.183 & 60.95 & 88.90 & 100.0 & 98.87 & 93.38 & 0.626 & 87.82  \\
AE4AD \cite{cai2024rethinking} & 70.83 & 80.75 & 67.26 & 71.27 & 0.291 & 67.38 & 76.09 & 65.75 & 68.55 & 70.68 & 0.264 & 65.96  \\
MMAD-4o & 70.00 & 85.00 & 100.0 & 80.59 & 0.726 & 82.23 & 93.75 & 100.0 & 100.0 & 96.48 & 0.832 & 93.32  \\
MMAD-4o-mini & 80.00 & 80.00 & 97.50 & 84.12 & 0.665 & 80.40 & 92.75 & 96.25 & 99.13 & 95.21 & 0.756 & 89.79  \\
MMAD-Sonnet & 52.22 & 75.00 & 82.50 & 64.71 & 0.537 & 67.75 & 87.32 & 97.75 & 99.61 & 92.48 & 0.756 & 89.44  \\
MMAD-Haiku & 50.00 & 50.00 & 50.00 & 50.00 & nan & 50.00 & 83.75 & 100.0 & 100.0 & 90.85 & 0.748 & 85.89 \\
\hline
        & \multicolumn{6}{c|}{\textbf{fryum}} 
        & \multicolumn{6}{c}{\textbf{macaroni1}}\\
\cline{1-13}
Skip-GAN \cite{akccay2019skip} & 49.00 & 92.91 & 100.0 & 94.46 & 0.680 & 91.15 & 96.06 & 90.00 & 100.0 & 97.10 & 0.682 & 95.34 \\
RD4AD  \cite{deng2022anomaly} & 73.33 & 90.51 & 99.97 & 95.06 & 0.746 & 95.15 & 97.38 & 97.33 & 97.63 & 97.51 & 0.664 & 94.11 \\
PatchCore \cite{roth2022towards} & 78.67 & 92.63 & 100.0 & 96.14 & 0.758 & 95.86 & 97.19 & 98.58 & 96.66 & 97.16 & 0.628 & 91.74 \\
CFLOW-AD \cite{gudovskiy2022cflow} & 60.67 & 83.77 & 96.98 & 90.18 & 0.632 & 88.27 & 87.38 & 91.42 & 77.14 & 82.46 & 0.397 & 76.38 \\
RRD \cite{tien2023revisiting} & 77.33 & 86.29 & 99.86 & 93.76 & 0.731 & 94.21 & 91.31 & 94.75 & 94.23 & 93.59 & 0.597 & 89.69 \\
OCR-GAN \cite{liang2023omni} & 94.00 & 86.57 & 88.61 & 88.22 & 0.454 & 77.45 & 98.19 & 98.75 & 93.37 & 95.62 & 0.633 & 92.04 \\
PNI \cite{bae2023pni} & 85.00 & 97.89 & 100.0 & 98.36 & 0.759 & 95.92 & 97.56 & 99.42 & 96.46 & 97.30 & 0.605 & 90.22 \\
SPR \cite{shin2023anomaly} & 60.00 & 62.97 & 75.73 & 70.32 & 0.305 & 68.44 & 52.50 & 47.50 & 71.34 & 62.02 & 0.193 & 62.81 \\
IGD \cite{chen2022deep} & 82.33 & 77.49 & 90.81 & 85.64 & 0.516 & 81.24 & 80.50 & 78.25 & 65.43 & 71.70 & 0.241 & 66.03 \\
AE4AD \cite{cai2024rethinking} & 87.69 & 63.37 & 85.67 & 79.06 & 0.436 & 76.39 & 64.86 & 79.58 & 82.06 & 72.03 & 0.248 & 66.50 \\
MMAD-4o & 50.00 & 77.14 & 96.61 & 87.00 & 0.745 & 87.86 & 52.62 & 57.83 & 82.40 & 70.16 & 0.584 & 70.80 \\
MMAD-4o-mini & 54.33 & 75.26 & 94.68 & 85.46 & 0.688 & 85.12 & 49.62 & 63.17 & 73.31 & 65.37 & 0.423 & 65.87 \\
MMAD-Sonnet & 50.00 & 70.00 & 89.83 & 80.50 & 0.665 & 81.37 & 56.25 & 58.33 & 82.86 & 71.43 & 0.605 & 71.92 \\
MMAD-Haiku & 50.00 & 55.71 & 67.80 & 62.50 & 0.382 & 62.64 & 49.00 & 49.00 & 69.20 & 60.22 & 0.396 & 61.24 \\
\hline
        & \multicolumn{6}{c|}{\textbf{macaroni2}} 
        & \multicolumn{6}{c}{\textbf{pcb1}}\\
\cline{1-13}
Skip-GAN \cite{akccay2019skip} & 45.00 & 53.67 & 50.46 & 49.48 & 0.000 & 49.99 & 82.00 & 87.24 & 90.05 & 87.07 & 0.513 & 82.35 \\
RD4AD  \cite{deng2022anomaly} & 91.81 & 80.56 & 87.40 & 87.55 & 0.479 & 82.21 & 97.71 & 97.02 & 94.20 & 96.55 & 0.604 & 88.08 \\
PatchCore \cite{roth2022towards} & 91.56 & 77.33 & 69.77 & 76.72 & 0.300 & 70.21 & 98.50 & 98.61 & 96.70 & 98.21 & 0.644 & 90.57 \\
CFLOW-AD \cite{gudovskiy2022cflow} & 58.50 & 46.78 & 50.94 & 52.33 & 0.021 & 51.39 & 94.86 & 93.70 & 99.05 & 94.93 & 0.646 & 90.68 \\
RRD \cite{tien2023revisiting} & 92.56 & 71.67 & 85.69 & 85.42 & 0.447 & 80.08 & 94.29 & 95.50 & 94.70 & 95.17 & 0.612 & 88.54 \\
OCR-GAN \cite{liang2023omni} & 94.00 & 90.56 & 93.00 & 92.90 & 0.545 & 86.66 & 93.93 & 93.56 & 93.65 & 93.63 & 0.569 & 85.87 \\
PNI \cite{bae2023pni} & 98.19 & 79.56 & 77.26 & 83.18 & 0.364 & 74.51 & 99.79 & 99.61 & 98.90 & 99.49 & 0.656 & 91.36 \\
SPR \cite{shin2023anomaly} & 55.56 & 58.33 & 53.80 & 54.95 & 0.059 & 53.97 & 30.14 & 54.29 & 41.60 & 48.37 & 0.014 & 49.14 \\
IGD \cite{chen2022deep} & 66.75 & 70.00 & 62.46 & 64.73 & 0.170 & 61.43 & 89.71 & 90.20 & 98.85 & 91.86 & 0.618 & 88.92 \\
AE4AD \cite{cai2024rethinking} & 75.26 & 73.22 & 79.25 & 76.02 & 0.320 & 71.53 & 84.82 & 95.40 & 90.64 & 87.75 & 0.522 & 82.87 \\
MMAD-4o & 53.12 & 50.00 & 61.43 & 57.50 & 0.342 & 57.52 & 58.82 & 81.01 & 71.40 & 75.98 & 0.536 & 73.38 \\
MMAD-4o-mini & 38.50 & 60.61 & 60.13 & 54.43 & 0.108 & 55.93 & 60.86 & 75.00 & 70.10 & 72.04 & 0.448 & 69.26 \\
MMAD-Sonnet & 51.28 & 47.78 & 60.00 & 55.84 & 0.120 & 56.20 & 53.36 & 76.83 & 90.95 & 76.36 & 0.585 & 76.58 \\
MMAD-Haiku & 48.00 & 53.56 & 50.86 & 50.50 & 0.033 & 50.65 & 49.50 & 51.03 & 57.07 & 52.02 & 0.166 & 52.54 \\
    \bottomrule
    
    \end{tabular}}
    \caption{Full results on VisA-MAD (Part I)}
    \label{tab:visa_detailed1}
\end{table*}

\begin{table*}[htbp]\scriptsize
    \centering
    \renewcommand{\arraystretch}{1.2}
    \scalebox{1}{
\setlength{\tabcolsep}{2.85pt} % Adjust column separation
    \begin{tabular}{p{2cm}cccccc|cccccc}
    \toprule
        \multirow{3}{*}{\textbf{Method}} 
        & \multicolumn{4}{c}{\textbf{Binary AD performance}} 
        & \multicolumn{2}{c|}{\textbf{MAD performance}}
        & \multicolumn{4}{c}{\textbf{Binary AD performance}}
        & \multicolumn{2}{c}{\textbf{MAD performance}}\\
\cline{2-13}

        & \textbf{Level 1} & \textbf{Level 2} & \textbf{ Level 3} 
        & \textbf{Whole} & \textbf{\;Ken} & \textbf{C} 
        & \textbf{Level 1} & \textbf{Level 2} & \textbf{ Level 3} 
        & \textbf{Whole} & \textbf{\;Ken} & \textbf{C} \\
\cline{2-13}
\hline
        & \multicolumn{6}{c|}{\textbf{pcb2}} 
        & \multicolumn{6}{c}{\textbf{pcb3}}\\
\cline{1-13}
Skip-GAN \cite{akccay2019skip} & 75.57 & 99.37 & 100.0 & 96.16 & 0.703 & 94.41 & 71.68 & 93.71 & 94.50 & 89.47 & 0.576 & 85.96 \\
RD4AD  \cite{deng2022anomaly} & 98.93 & 97.67 & 90.00 & 96.39 & 0.555 & 85.04 & 96.68 & 98.00 & 90.10 & 96.16 & 0.570 & 85.57 \\
PatchCore \cite{roth2022towards} & 100.0 & 98.46 & 91.84 & 97.42 & 0.578 & 86.48 & 97.82 & 99.08 & 97.08 & 98.43 & 0.599 & 87.38 \\
CFLOW-AD \cite{gudovskiy2022cflow} & 92.43 & 89.75 & 86.47 & 89.50 & 0.516 & 82.57 & 85.20 & 84.85 & 85.94 & 85.14 & 0.458 & 78.61 \\
RRD \cite{tien2023revisiting} & 98.36 & 93.04 & 78.95 & 91.11 & 0.484 & 80.56 & 96.73 & 96.40 & 94.46 & 96.08 & 0.590 & 86.82 \\
OCR-GAN \cite{liang2023omni} & 95.71 & 96.85 & 98.79 & 97.06 & 0.666 & 92.07 & 82.97 & 85.59 & 88.76 & 85.70 & 0.504 & 81.48 \\
PNI \cite{bae2023pni} & 100.0 & 99.75 & 95.84 & 99.04 & 0.596 & 87.66 & 99.36 & 99.79 & 99.01 & 99.54 & 0.609 & 88.03 \\
SPR \cite{shin2023anomaly} & 48.29 & 58.40 & 47.58 & 54.93 & 0.058 & 53.64 & 51.78 & 63.20 & 68.66 & 62.01 & 0.184 & 61.48 \\
IGD \cite{chen2022deep} & 71.93 & 83.16 & 93.63 & 83.58 & 0.486 & 80.70 & 71.88 & 76.65 & 88.47 & 78.06 & 0.414 & 75.81 \\
AE4AD \cite{cai2024rethinking} & 96.66 & 99.53 & 90.29 & 96.31 & 0.683 & 93.10 & 74.26 & 90.54 & 71.73 & 77.01 & 0.407 & 75.39 \\
MMAD-4o & 9.89 & 75.96 & 71.84 & 75.72 & 0.490 & 71.94 & 67.50 & 78.33 & 62.50 & 73.00 & 0.475 & 68.62 \\
MMAD-4o-mini & 70.00 & 69.93 & 62.68 & 68.56 & 0.290 & 64.06 & 63.24 & 80.09 & 59.95 & 72.69 & 0.367 & 67.36 \\
MMAD-Sonnet & 77.71 & 81.66 & 73.16 & 79.49 & 0.556 & 74.74 & 67.50 & 72.50 & 67.50 & 70.50 & 0.450 & 66.78 \\
MMAD-Haiku & 62.89 & 65.05 & 51.21 & 62.12 & 0.257 & 58.36 & 51.98 & 56.20 & 49.50 & 54.02 & 0.148 & 52.87 \\
        & \multicolumn{6}{c|}{\textbf{pipe\_fryum}} 
        & \multicolumn{6}{c}{\textbf{mean}}\\
\cline{1-13}
Skip-GAN \cite{akccay2019skip} & 63.69 & 73.07 & 88.00 & 73.16 & 0.380 & 72.49 & 69.85 & 84.62 & 87.79 & 82.64 & 0.487 & 80.29 \\
RD4AD  \cite{deng2022anomaly} & 98.41 & 100.0 & 99.84 & 99.16 & 0.701 & 91.54 & 92.76 & 95.49 & 95.29 & 95.24 & 0.640 & 89.79 \\
PatchCore \cite{roth2022towards} & 99.38 & 100.0 & 100.0 & 99.70 & 0.708 & 91.91 & 91.89 & 94.46 & 94.09 & 93.32 & 0.602 & 87.42 \\
CFLOW-AD \cite{gudovskiy2022cflow} & 95.38 & 100.0 & 97.92 & 97.06 & 0.640 & 87.92 & 81.54 & 87.14 & 86.89 & 85.08 & 0.493 & 80.43 \\
RRD \cite{tien2023revisiting} & 99.38 & 99.87 & 99.92 & 99.65 & 0.688 & 90.72 & 92.65 & 93.01 & 94.15 & 93.99 & 0.620 & 88.53 \\
OCR-GAN \cite{liang2023omni} & 80.97 & 97.60 & 94.32 & 88.35 & 0.483 & 78.61 & 91.23 & 91.86 & 91.73 & 91.17 & 0.526 & 83.03 \\
PNI \cite{bae2023pni} & 99.74 & 100.0 & 99.76 & 99.80 & 0.662 & 89.21 & 95.54 & 96.77 & 96.20 & 96.07 & 0.622 & 88.64 \\
SPR \cite{shin2023anomaly} & 49.49 & 46.53 & 37.44 & 45.11 & 0.097 & 44.23 & 50.10 & 56.45 & 55.65 & 56.05 & 0.087 & 55.49 \\
IGD \cite{chen2022deep} & 75.64 & 90.40 & 92.40 & 83.75 & 0.516 & 80.56 & 75.98 & 81.36 & 84.34 & 79.34 & 0.419 & 75.94 \\
AE4AD \cite{cai2024rethinking} & 62.40 & 63.60 & 45.79 & 54.58 & 0.115 & 56.79 & 76.99 & 79.08 & 75.69 & 76.08 & 0.365 & 72.88 \\
MMAD-4o & 96.15 & 98.47 & 99.56 & 97.67 & 0.857 & 94.03 & 61.32 & 78.19 & 82.86 & 79.34 & 0.621 & 77.74 \\
MMAD-4o-mini & 85.18 & 87.87 & 98.96 & 90.05 & 0.756 & 88.76 & 66.05 & 76.46 & 79.60 & 76.44 & 0.500 & 74.06 \\
MMAD-Sonnet & 92.00 & 96.60 & 98.36 & 94.89 & 0.775 & 88.97 & 65.29 & 75.16 & 82.75 & 76.24 & 0.561 & 74.86 \\
MMAD-Haiku & 69.23 & 93.33 & 70.00 & 74.05 & 0.479 & 69.69 & 57.15 & 63.76 & 62.85 & 61.81 & 0.326 & 60.43 \\
    \bottomrule
    
    \end{tabular}}
    \caption{Full results on VisA-MAD (Part II)}
    \label{tab:visa_detailed2}
\end{table*}

%% file: sec/tables/mvtec_detailed.tex
\begin{table*}[htbp]\scriptsize
    \centering
    \renewcommand{\arraystretch}{1.2}
    \scalebox{1}{
\setlength{\tabcolsep}{2.85pt} % Adjust column separation
    \begin{tabular}{p{2cm}cccccc|cccccc}
    \toprule
        \multirow{3}{*}{\textbf{Method}} 
        & \multicolumn{4}{c}{\textbf{Binary AD performance}} 
        & \multicolumn{2}{c|}{\textbf{MAD performance}}
        & \multicolumn{4}{c}{\textbf{Binary AD performance}}
        & \multicolumn{2}{c}{\textbf{MAD performance}}\\
\cline{2-13}

        & \textbf{Level 1} & \textbf{Level 2} & \textbf{ Level 3} 
        & \textbf{Whole} & \textbf{\;Ken} & \textbf{C} 
        & \textbf{Level 1} & \textbf{Level 2} & \textbf{ Level 3} 
        & \textbf{Whole} & \textbf{\;Ken} & \textbf{C} \\
\cline{2-13}

        & \multicolumn{6}{c|}{\textbf{carpet}} 
        & \multicolumn{6}{c}{\textbf{grid}}\\
\cline{1-13}
    % Example row, replace with actual data
    % \hline
Skip-GAN \cite{akccay2019skip} & 29.20 & 47.18 & 70.90 & 54.34 & 0.243 & 64.13 & 72.87 & 90.08 & 54.37 & 72.60 & 0.150 & 58.88 \\
RD4AD  \cite{deng2022anomaly} & 100.0 & 100.0 & 100.0 & 100.0 & 0.587 & 84.17 & 100.0 & 100.0 & 100.0 & 100.0 & 0.616 & 86.54 \\
PatchCore \cite{roth2022towards} & 97.06 & 100.0 & 100.0 & 99.29 & 0.663 & 88.59 & 96.54 & 100.0 & 100.0 & 97.99 & 0.659 & 89.08 \\
CFLOW-AD \cite{gudovskiy2022cflow} & 100.0 & 93.61 & 100.0 & 98.27 & 0.446 & 75.93 & 90.91 & 94.84 & 98.81 & 93.40 & 0.482 & 78.62 \\
RRD \cite{tien2023revisiting} & 100.0 & 100.0 & 100.0 & 100.0 & 0.582 & 83.89 & 100.0 & 100.0 & 100.0 & 100.0 & 0.659 & 89.08 \\
OCR-GAN \cite{liang2023omni} & 94.12 & 100.0 & 100.0 & 98.57 & 0.286 & 66.67 & 97.40 & 100.0 & 100.0 & 98.50 & 0.550 & 82.61 \\
PNI \cite{bae2023pni} & 100.0 & 99.44 & 100.0 & 99.85 & 0.616 & 85.83 & 97.40 & 100.0 & 100.0 & 98.50 & 0.579 & 84.34 \\
SPR \cite{shin2023anomaly} & 83.40 & 100.0 & 95.06 & 93.57 & 0.443 & 75.79 & 100.0 & 100.0 & 100.0 & 100.0 & 0.342 & 70.28 \\
IGD \cite{chen2022deep} & 64.29 & 73.87 & 86.34 & 77.60 & 0.383 & 72.28 & 68.25 & 72.62 & 59.52 & 67.34 & 0.127 & 57.52 \\
AE4AD \cite{cai2024rethinking} & 5.46 & 78.20 & 60.50 & 51.94 & 0.171 & 59.94 & 66.81 & 82.54 & 69.84 & 70.76 & 0.268 & 65.87 \\
MMAD-4o & 100.0 & 100.0 & 100.0 & 100.0 & 0.811 & 92.10 & 98.48 & 100.0 & 100.0 & 99.12 & 0.683 & 86.80 \\
MMAD-4o-mini & 100.0 & 100.0 & 100.0 & 100.0 & 0.752 & 87.62 & 94.23 & 98.81 & 100.0 & 96.41 & 0.755 & 89.52 \\
MMAD-Sonnet & 100.0 & 100.0 & 98.53 & 99.29 & 0.744 & 86.93 & 96.97 & 95.83 & 100.0 & 97.37 & 0.728 & 86.97 \\
MMAD-Haiku & 91.18 & 89.47 & 94.12 & 92.14 & 0.579 & 78.81 & 80.30 & 83.33 & 100.0 & 85.09 & 0.552 & 75.55 \\
\hline
        & \multicolumn{6}{c|}{\textbf{leather}} 
        & \multicolumn{6}{c}{\textbf{tile}}\\
\cline{1-13}
Skip-GAN \cite{akccay2019skip} & 47.70 & 16.36 & 10.05 & 26.77 & 0.437 & 24.55 & 76.68 & 74.68 & 85.92 & 77.81 & 0.249 & 64.45 \\
RD4AD  \cite{deng2022anomaly} & 100.0 & 100.0 & 100.0 & 100.0 & 0.473 & 77.55 & 100.0 & 98.44 & 100.0 & 99.42 & 0.422 & 74.52 \\
PatchCore \cite{roth2022towards} & 100.0 & 100.0 & 100.0 & 100.0 & 0.456 & 76.55 & 100.0 & 96.97 & 100.0 & 98.88 & 0.445 & 75.85 \\
CFLOW-AD \cite{gudovskiy2022cflow} & 99.84 & 100.0 & 100.0 & 99.93 & 0.412 & 73.96 & 100.0 & 98.34 & 100.0 & 99.39 & 0.391 & 72.73 \\
RRD \cite{tien2023revisiting} & 100.0 & 100.0 & 100.0 & 100.0 & 0.536 & 81.21 & 100.0 & 99.22 & 100.0 & 99.71 & 0.440 & 75.57 \\
OCR-GAN \cite{liang2023omni} & 100.0 & 100.0 & 100.0 & 100.0 & 0.654 & 88.05 & 99.49 & 96.19 & 96.43 & 97.66 & 0.247 & 64.35 \\
PNI \cite{bae2023pni} & 99.42 & 100.0 & 100.0 & 99.76 & 0.464 & 77.01 & 100.0 & 99.61 & 100.0 & 99.86 & 0.461 & 76.77 \\
SPR \cite{shin2023anomaly} & 99.84 & 100.0 & 99.92 & 99.90 & 0.425 & 74.72 & 99.66 & 100.0 & 100.0 & 99.86 & 0.398 & 73.11 \\
IGD \cite{chen2022deep} & 88.49 & 97.24 & 87.58 & 89.74 & 0.363 & 71.11 & 100.0 & 93.45 & 100.0 & 97.58 & 0.469 & 77.26 \\
AE4AD \cite{cai2024rethinking} & 67.68 & 97.43 & 73.90 & 75.68 & 0.290 & 66.90 & 51.52 & 54.25 & 57.22 & 53.68 & 0.055 & 53.19 \\
MMAD-4o & 99.05 & 100.0 & 100.0 & 99.61 & 0.782 & 91.27 & 100.0 & 100.0 & 100.0 & 100.0 & 0.623 & 82.74 \\
MMAD-4o-mini & 98.89 & 100.0 & 100.0 & 99.54 & 0.750 & 86.30 & 100.0 & 87.10 & 100.0 & 95.24 & 0.483 & 75.17 \\
MMAD-Sonnet & 100.0 & 100.0 & 100.0 & 100.0 & 0.658 & 82.28 & 100.0 & 95.01 & 100.0 & 98.16 & 0.544 & 78.06 \\
MMAD-Haiku & 98.48 & 99.91 & 98.44 & 98.73 & 0.578 & 78.38 & 94.44 & 74.19 & 100.0 & 88.10 & 0.477 & 73.71 \\
\hline
        & \multicolumn{6}{c|}{\textbf{wood}} 
        & \multicolumn{6}{c}{\textbf{bottle}}\\
\cline{1-13}
Skip-GAN \cite{akccay2019skip} & 83.92 & 95.76 & 89.00 & 90.96 & 0.451 & 76.49 & 60.00 & 58.41 & 42.00 & 53.73 & 0.078 & 45.55 \\
RD4AD  \cite{deng2022anomaly} & 100.0 & 98.47 & 100.0 & 99.21 & 0.548 & 82.16 & 99.76 & 100.0 & 100.0 & 99.92 & 0.602 & 84.55 \\
PatchCore \cite{roth2022towards} & 100.0 & 98.13 & 100.0 & 99.04 & 0.491 & 78.81 & 100.0 & 100.0 & 100.0 & 100.0 & 0.580 & 83.27 \\
CFLOW-AD \cite{gudovskiy2022cflow} & 100.0 & 97.28 & 100.0 & 98.60 & 0.474 & 77.78 & 100.0 & 100.0 & 100.0 & 100.0 & 0.526 & 80.21 \\
RRD \cite{tien2023revisiting} & 100.0 & 98.64 & 100.0 & 99.30 & 0.535 & 81.40 & 100.0 & 100.0 & 100.0 & 100.0 & 0.567 & 82.57 \\
OCR-GAN \cite{liang2023omni} & 87.43 & 100.0 & 100.0 & 96.23 & 0.688 & 90.34 & 97.14 & 100.0 & 100.0 & 99.05 & 0.458 & 76.30 \\
PNI \cite{bae2023pni} & 100.0 & 98.47 & 100.0 & 99.21 & 0.438 & 75.68 & 100.0 & 100.0 & 100.0 & 100.0 & 0.609 & 84.93 \\
SPR \cite{shin2023anomaly} & 99.84 & 100.0 & 99.92 & 99.90 & 0.425 & 74.72 & 99.66 & 100.0 & 100.0 & 99.86 & 0.398 & 73.11 \\
IGD \cite{chen2022deep} & 100.0 & 95.93 & 100.0 & 97.89 & 0.461 & 77.02 & 98.81 & 99.77 & 100.0 & 99.52 & 0.453 & 75.99 \\
AE4AD \cite{cai2024rethinking} & 85.67 & 86.08 & 82.78 & 85.35 & 0.372 & 71.84 & 95.48 & 93.18 & 100.0 & 96.11 & 0.441 & 75.33 \\
MMAD-4o & 94.88 & 95.93 & 98.33 & 96.05 & 0.596 & 81.31 & 100.0 & 100.0 & 100.0 & 100.0 & 0.721 & 87.66 \\
MMAD-4o-mini & 95.18 & 97.45 & 99.28 & 97.11 & 0.628 & 83.57 & 98.57 & 100.0 & 100.0 & 99.52 & 0.731 & 89.78 \\
MMAD-Sonnet & 100.0 & 99.49 & 100.0 & 99.74 & 0.580 & 79.73 & 86.67 & 99.55 & 100.0 & 95.40 & 0.639 & 80.89 \\
MMAD-Haiku & 96.78 & 95.67 & 97.13 & 96.27 & 0.579 & 75.57 & 76.19 & 90.91 & 92.50 & 86.51 & 0.607 & 79.45 \\
\hline
        & \multicolumn{6}{c|}{\textbf{cable}} 
        & \multicolumn{6}{c}{\textbf{capsule}}\\
\cline{1-13}
Skip-GAN \cite{akccay2019skip} & 47.75 & 32.52 & 24.45 & 37.19 & 0.219 & 36.55 & 44.83 & 94.57 & 50.59 & 56.28 & 0.078 & 54.56 \\
RD4AD  \cite{deng2022anomaly} & 95.73 & 99.14 & 99.22 & 97.75 & 0.568 & 84.98 & 95.85 & 100.0 & 98.32 & 97.61 & 0.543 & 81.89 \\
PatchCore \cite{roth2022towards} & 99.78 & 99.37 & 100.0 & 99.66 & 0.635 & 89.10 & 96.14 & 100.0 & 98.81 & 97.93 & 0.480 & 78.17 \\
CFLOW-AD \cite{gudovskiy2022cflow} & 94.60 & 95.61 & 99.69 & 96.00 & 0.579 & 85.62 & 92.08 & 99.78 & 97.73 & 95.77 & 0.437 & 75.65 \\
RRD \cite{tien2023revisiting} & 99.55 & 100.0 & 100.0 & 99.82 & 0.591 & 86.37 & 98.26 & 100.0 & 99.21 & 98.96 & 0.545 & 82.00 \\
OCR-GAN \cite{liang2023omni} & 78.94 & 94.12 & 88.87 & 86.85 & 0.524 & 82.28 & 90.05 & 100.0 & 95.95 & 94.26 & 0.270 & 65.87 \\
PNI \cite{bae2023pni} & 98.35 & 98.90 & 100.0 & 98.89 & 0.676 & 91.60 & 99.42 & 100.0 & 99.31 & 99.48 & 0.360 & 71.17 \\
SPR \cite{shin2023anomaly} & 97.90 & 85.97 & 99.53 & 93.53 & 0.527 & 82.47 & 95.94 & 100.0 & 97.83 & 97.45 & 0.581 & 84.12 \\
IGD \cite{chen2022deep} & 90.93 & 93.18 & 98.28 & 93.26 & 0.572 & 85.22 & 80.19 & 99.78 & 86.46 & 86.32 & 0.347 & 70.40 \\
AE4AD \cite{cai2024rethinking} & 64.77 & 88.56 & 87.77 & 78.63 & 0.435 & 76.75 & 62.80 & 99.35 & 63.44 & 69.76 & 0.150 & 58.78 \\
MMAD-4o & 99.06 & 99.96 & 100.0 & 99.60 & 0.835 & 93.28 & 91.88 & 94.67 & 96.44 & 94.24 & 0.422 & 72.43 \\
MMAD-4o-mini & 93.85 & 94.59 & 94.20 & 94.21 & 0.647 & 84.93 & 88.02 & 99.57 & 95.16 & 93.02 & 0.492 & 74.01 \\
MMAD-Sonnet & 94.19 & 97.02 & 98.98 & 96.24 & 0.805 & 88.66 & 82.85 & 81.52 & 86.41 & 84.04 & 0.356 & 67.59 \\
MMAD-Haiku & 60.87 & 75.00 & 81.82 & 70.54 & 0.519 & 68.75 & 61.11 & 50.00 & 70.45 & 62.84 & 0.269 & 59.59 \\
    \bottomrule
    
    \end{tabular}}
    \caption{Full results on MVTec-MAD (Part I)}
    \label{tab:mvtec_detailed1}
\end{table*}

\begin{table*}[htbp]\scriptsize
    \centering
    \renewcommand{\arraystretch}{1.2}
    \scalebox{1}{
\setlength{\tabcolsep}{2.85pt} % Adjust column separation
    \begin{tabular}{p{2cm}cccccc|cccccc}
    \toprule
        \multirow{3}{*}{\textbf{Method}} 
        & \multicolumn{4}{c}{\textbf{Binary AD performance}} 
        & \multicolumn{2}{c|}{\textbf{MAD performance}}
        & \multicolumn{4}{c}{\textbf{Binary AD performance}}
        & \multicolumn{2}{c}{\textbf{MAD performance}}\\
\cline{2-13}

        & \textbf{Level 1} & \textbf{Level 2} & \textbf{ Level 3} 
        & \textbf{Whole} & \textbf{\;Ken} & \textbf{C} 
        & \textbf{Level 1} & \textbf{Level 2} & \textbf{ Level 3} 
        & \textbf{Whole} & \textbf{\;Ken} & \textbf{C} \\
\cline{2-13}

        & \multicolumn{6}{c|}{\textbf{hazelnut}} 
        & \multicolumn{6}{c}{\textbf{metal\_nut}}\\
\cline{1-13}
Skip-GAN \cite{akccay2019skip} & 100.0 & 69.56 & 77.57 & 81.07 & 0.218 & 62.83 & 43.39 & 26.09 & 57.27 & 42.66 & 0.084 & 54.84 \\
RD4AD  \cite{deng2022anomaly} & 100.0 & 100.0 & 100.0 & 100.0 & 0.560 & 82.98 & 100.0 & 100.0 & 100.0 & 100.0 & 0.469 & 76.92 \\
PatchCore \cite{roth2022towards} & 100.0 & 100.0 & 100.0 & 100.0 & 0.504 & 79.71 & 100.0 & 99.60 & 100.0 & 99.87 & 0.455 & 76.13 \\
CFLOW-AD \cite{gudovskiy2022cflow} & 100.0 & 100.0 & 100.0 & 100.0 & 0.611 & 86.02 & 98.76 & 90.91 & 99.64 & 96.49 & 0.430 & 74.68 \\
RRD \cite{tien2023revisiting} & 100.0 & 100.0 & 100.0 & 100.0 & 0.563 & 83.19 & 100.0 & 100.0 & 100.0 & 100.0 & 0.504 & 78.93 \\
OCR-GAN \cite{liang2023omni} & 99.85 & 90.15 & 99.10 & 97.11 & 0.482 & 78.44 & 73.55 & 100.0 & 99.09 & 91.36 & 0.212 & 62.16 \\
PNI \cite{bae2023pni} & 100.0 & 100.0 & 100.0 & 100.0 & 0.448 & 76.42 & 100.0 & 100.0 & 100.0 & 100.0 & 0.431 & 74.74 \\
SPR \cite{shin2023anomaly} & 100.0 & 94.26 & 100.0 & 98.61 & 0.466 & 77.49 & 99.59 & 86.36 & 100.0 & 95.39 & 0.388 & 72.31 \\
IGD \cite{chen2022deep} & 93.38 & 95.29 & 98.40 & 96.43 & 0.636 & 87.48 & 79.55 & 94.86 & 75.45 & 83.12 & 0.222 & 62.72 \\
AE4AD \cite{cai2024rethinking} & 99.26 & 68.82 & 85.28 & 84.68 & 0.319 & 68.82 & 48.55 & 45.45 & 49.64 & 47.92 & 0.026 & 48.53 \\
MMAD-4o & 99.26 & 98.38 & 100.0 & 99.43 & 0.834 & 94.05 & 100.0 & 100.0 & 100.0 & 100.0 & 0.636 & 82.89 \\
MMAD-4o-mini & 85.07 & 84.19 & 98.58 & 91.80 & 0.713 & 88.72 & 89.15 & 99.01 & 91.73 & 93.31 & 0.636 & 82.73 \\
MMAD-Sonnet & 94.41 & 97.57 & 98.75 & 97.41 & 0.825 & 91.27 & 95.45 & 80.43 & 78.00 & 84.29 & 0.363 & 67.19 \\
MMAD-Haiku & 80.15 & 72.57 & 92.64 & 84.73 & 0.545 & 79.32 & 88.64 & 63.04 & 56.00 & 68.57 & 0.044 & 48.28 \\
\hline
        & \multicolumn{6}{c|}{\textbf{pill}} 
        & \multicolumn{6}{c}{\textbf{screw}}\\
\cline{1-13}
Skip-GAN \cite{akccay2019skip} & 66.55 & 80.33 & 96.83 & 76.70 & 0.512 & 80.01 & 65.85 & 0.00 & 0.00 & 22.55 & 0.496 & 21.20 \\
RD4AD  \cite{deng2022anomaly} & 100.0 & 92.46 & 100.0 & 97.72 & 0.438 & 75.64 & 99.41 & 99.59 & 95.93 & 98.33 & 0.286 & 66.60 \\
PatchCore \cite{roth2022towards} & 98.03 & 90.24 & 99.55 & 95.97 & 0.483 & 78.29 & 99.90 & 99.59 & 98.98 & 99.50 & 0.318 & 68.48 \\
CFLOW-AD \cite{gudovskiy2022cflow} & 94.19 & 85.36 & 96.83 & 92.04 & 0.457 & 76.78 & 97.07 & 97.15 & 95.02 & 96.42 & 0.439 & 75.51 \\
RRD \cite{tien2023revisiting} & 100.0 & 96.15 & 100.0 & 98.84 & 0.440 & 75.77 & 99.61 & 100.0 & 98.78 & 99.47 & 0.286 & 66.60 \\
OCR-GAN \cite{liang2023omni} & 97.50 & 100.0 & 98.19 & 98.39 & 0.185 & 60.86 & 84.68 & 50.61 & 74.29 & 70.06 & 0.122 & 57.10 \\
PNI \cite{bae2023pni} & 98.39 & 88.17 & 99.77 & 95.57 & 0.421 & 74.69 & 100.0 & 100.0 & 98.48 & 99.50 & 0.322 & 68.71 \\
SPR \cite{shin2023anomaly} & 92.49 & 89.94 & 100.0 & 93.20 & 0.495 & 78.97 & 98.73 & 98.98 & 98.88 & 98.86 & 0.420 & 74.40 \\
IGD \cite{chen2022deep} & 89.18 & 77.37 & 93.67 & 86.49 & 0.407 & 73.82 & 75.32 & 72.76 & 60.47 & 69.60 & 0.128 & 57.43 \\
AE4AD \cite{cai2024rethinking} & 67.62 & 81.07 & 96.61 & 77.42 & 0.476 & 77.91 & 27.90 & 40.96 & 61.38 & 43.20 & 0.055 & 53.20 \\
MMAD-4o & 97.67 & 84.62 & 100.0 & 94.19 & 0.498 & 75.68 & 96.00 & 77.08 & 81.25 & 84.93 & 0.480 & 72.44 \\
MMAD-4o-mini & 96.87 & 88.61 & 100.0 & 94.99 & 0.603 & 80.87 & 82.59 & 80.39 & 91.41 & 84.76 & 0.574 & 78.97 \\
MMAD-Sonnet & 75.22 & 59.69 & 87.44 & 72.94 & 0.307 & 63.63 & 72.39 & 65.75 & 74.44 & 70.88 & 0.323 & 65.36 \\
MMAD-Haiku & 60.87 & 50.00 & 59.73 & 57.36 & 0.054 & 51.63 & 54.00 & 52.08 & 52.08 & 52.74 & 0.080 & 51.22 \\
\hline
        & \multicolumn{6}{c|}{\textbf{transistor}} 
        & \multicolumn{6}{c}{\textbf{zipper}}\\
\cline{1-13}
Skip-GAN \cite{akccay2019skip} & 86.33 & 73.00 & 52.08 & 65.88 & 0.137 & 58.97 & 70.83 & 37.54 & 60.74 & 49.89 & 0.102 & 43.87 \\
RD4AD  \cite{deng2022anomaly} & 88.17 & 98.67 & 99.50 & 96.46 & 0.549 & 85.83 & 98.30 & 98.26 & 99.61 & 98.45 & 0.392 & 73.65 \\
PatchCore \cite{roth2022towards} & 100.0 & 100.0 & 100.0 & 100.0 & 0.625 & 90.83 & 99.15 & 99.51 & 100.0 & 99.47 & 0.368 & 72.20 \\
CFLOW-AD \cite{gudovskiy2022cflow} & 100.0 & 96.33 & 84.92 & 91.54 & 0.437 & 78.52 & 92.71 & 98.79 & 99.61 & 97.22 & 0.399 & 74.11 \\
RRD \cite{tien2023revisiting} & 95.83 & 99.67 & 99.17 & 98.46 & 0.570 & 87.24 & 97.73 & 98.66 & 99.80 & 98.56 & 0.434 & 76.19 \\
OCR-GAN \cite{liang2023omni} & 100.0 & 99.50 & 96.50 & 98.12 & 0.519 & 83.93 & 94.32 & 96.16 & 93.16 & 95.25 & 0.424 & 75.62 \\
PNI \cite{bae2023pni} & 100.0 & 100.0 & 100.0 & 100.0 & 0.573 & 87.41 & 99.62 & 99.87 & 100.0 & 99.82 & 0.421 & 75.41 \\
SPR \cite{shin2023anomaly} & 100.0 & 100.0 & 97.00 & 98.50 & 0.524 & 84.21 & 99.81 & 100.0 & 100.0 & 99.95 & 0.484 & 79.24 \\
IGD \cite{chen2022deep} & 100.0 & 88.83 & 86.67 & 90.54 & 0.426 & 77.83 & 87.03 & 92.81 & 98.05 & 91.91 & 0.383 & 73.11] \\
AE4AD \cite{cai2024rethinking} & 100.0 & 85.50 & 72.17 & 82.46 & 0.317 & 70.72 & 81.34 & 82.37 & 90.04 & 83.11 & 0.297 & 67.91 \\
MMAD-4o & 72.92 & 72.75 & 96.33 & 84.58 & 0.636 & 80.84 & 98.48 & 87.86 & 96.88 & 92.02 & 0.485 & 75.79 \\
MMAD-4o-mini & 67.67 & 65.67 & 87.00 & 76.83 & 0.478 & 75.16 & 98.48 & 87.14 & 96.88 & 91.60 & 0.358 & 69.09 \\
MMAD-Sonnet & 93.83 & 82.75 & 87.83 & 88.06 & 0.621 & 79.69 & 84.85 & 69.29 & 90.62 & 76.47 & 0.303 & 64.35 \\
MMAD-Haiku & 61.00 & 50.00 & 71.25 & 63.38 & 0.336 & 62.90 & 69.70 & 52.14 & 62.50 & 58.40 & 0.010 & 49.70 \\
    \bottomrule
    
    \end{tabular}}
    \caption{Full results on MVTec-MAD (Part II)}
    \label{tab:mvtec_detailed2}
\end{table*}

\begin{table*}[htbp]\scriptsize
    \centering
    \renewcommand{\arraystretch}{1.2}
    \scalebox{1}{
\setlength{\tabcolsep}{2.85pt} % Adjust column separation
    \begin{tabular}{p{2cm}cccccc}
    \toprule
        \multirow{3}{*}{\textbf{Method}} 
        & \multicolumn{4}{c}{\textbf{Binary AD performance}} 
        & \multicolumn{2}{c}{\textbf{MAD performance}}\\
\cline{2-7}

        & \textbf{Level 1} & \textbf{Level 2} & \textbf{ Level 3} 
        & \textbf{Whole} & \textbf{\;Ken} & \textbf{C} \\
\cline{2-7}
        & \multicolumn{6}{c}{\textbf{mean}} \\
\cline{1-7}
Skip-GAN \cite{akccay2019skip} & 63.99 & 56.86 & 55.13 & 57.75 & 0.057 & 53.35 \\
RD4AD  \cite{deng2022anomaly} & 98.37 & 98.93 & 99.47 & 98.92 & 0.504 & 79.86 \\
PatchCore \cite{roth2022towards} & 99.04 & 98.82 & 99.81 & 99.11 & 0.511 & 80.36 \\
CFLOW-AD \cite{gudovskiy2022cflow} & 97.15 & 96.29 & 98.02 & 96.79 & 0.466 & 77.58 \\
RRD \cite{tien2023revisiting} & 99.36 & 99.45 & 99.78 & 99.51 & 0.518 & 80.72 \\
OCR-GAN \cite{liang2023omni} & 92.46 & 94.77 & 95.83 & 94.39 & 0.402 & 73.90 \\
PNI \cite{bae2023pni} & 99.47 & 98.89 & 99.83 & 99.32 & 0.487 & 78.91 \\
SPR \cite{shin2023anomaly} & 97.63 & 96.82 & 99.15 & 97.76 & 0.451 & 76.78 \\
IGD \cite{chen2022deep} & 86.82 & 89.13 & 87.92 & 87.67 & 0.384 & 72.78 \\
AE4AD \cite{cai2024rethinking} & 66.06 & 77.41 & 75.04 & 71.48 & 0.259 & 65.41 \\
MMAD-4o & 96.26 & 93.66 & 97.80 & 95.98 & 0.646 & 83.52 \\
MMAD-4o-mini & 92.04 & 91.61 & 96.73 & 93.45 & 0.614 & 81.89 \\
MMAD-Sonnet & 91.20 & 87.42 & 92.93 & 90.02 & 0.557 & 77.33 \\
MMAD-Haiku & 76.69 & 71.31 & 80.62 & 76.10 & 0.366 & 66.63 \\
    \bottomrule
    
    \end{tabular}}
    \caption{Full results on MVTec-MAD (Part III)}
    \label{tab:mvtec_detailed3}
\end{table*}

%% file: sec/tables/multidogs_detailed.tex
\begin{table*}[htbp]\scriptsize
    \centering
    \renewcommand{\arraystretch}{1.2}
    \scalebox{1}{
\setlength{\tabcolsep}{2.85pt} % Adjust column separation
    \begin{tabular}{p{2cm}ccccccc|ccccccc}
    \toprule
        \multirow{3}{*}{\textbf{Method}} 
        & \multicolumn{5}{c}{\textbf{Binary AD performance}} 
        & \multicolumn{2}{c|}{\textbf{MAD performance}}
        & \multicolumn{5}{c}{\textbf{Binary AD performance}}
        & \multicolumn{2}{c}{\textbf{MAD performance}}\\
\cline{2-15}

        & \textbf{Level 1} & \textbf{Level 2} & \textbf{ Level 3} & \textbf{ Level 4}
        & \textbf{Whole} & \textbf{\;Ken} & \textbf{C} 
        & \textbf{Level 1} & \textbf{Level 2} & \textbf{ Level 3} & \textbf{ Level 4}
        & \textbf{Whole} & \textbf{\;Ken} & \textbf{C} \\
\cline{2-15}

        & \multicolumn{7}{c|}{\textbf{bichon\_frise}} 
        & \multicolumn{7}{c}{\textbf{chinese\_rural\_dog}}\\
\cline{1-15}
    % Example row, replace with actual data
    % \hline
Skip-GAN \cite{akccay2019skip} & 92.80 & 98.22 & 85.30 & 99.95 & 94.07 & 0.538 & 80.08 & 79.79 & 89.59 & 85.74 & 99.64 & 88.69 & 0.549 & 80.70 \\
RD4AD  \cite{deng2022anomaly} & 63.99 & 73.69 & 78.60 & 80.36 & 74.16 & 0.311 & 67.37 & 54.87 & 61.76 & 71.06 & 80.03 & 66.93 & 0.290 & 66.22 \\
PatchCore \cite{roth2022towards} & 66.82 & 81.22 & 79.02 & 87.93 & 78.74 & 0.408 & 72.78 & 53.73 & 74.47 & 74.92 & 88.68 & 72.95 & 0.396 & 72.15 \\
CFLOW-AD \cite{gudovskiy2022cflow} & 93.38 & 97.57 & 97.26 & 95.72 & 95.98 & 0.366 & 70.46 & 61.95 & 86.39 & 89.37 & 91.08 & 82.20 & 0.438 & 74.49 \\
RRD \cite{tien2023revisiting} & 75.90 & 87.30 & 90.52 & 96.01 & 87.43 & 0.523 & 79.21 & 65.95 & 73.51 & 85.10 & 97.59 & 80.54 & 0.509 & 78.47 \\
OCR-GAN \cite{liang2023omni} & 78.46 & 72.60 & 69.00 & 96.58 & 79.16 & 0.383 & 71.40 & 73.92 & 75.54 & 63.97 & 94.01 & 76.86 & 0.339 & 68.94 \\
PNI \cite{bae2023pni} & 71.16 & 82.80 & 78.79 & 89.51 & 80.57 & 0.413 & 73.07 & 64.60 & 77.04 & 75.50 & 90.32 & 76.87 & 0.402 & 72.47 \\
SPR \cite{shin2023anomaly} & 61.83 & 55.24 & 79.30 & 71.25 & 66.90 & 0.238 & 63.30 & 59.69 & 54.78 & 80.94 & 72.90 & 67.08 & 0.249 & 63.92 \\
IGD \cite{chen2022deep} & 97.67 & 97.38 & 96.44 & 96.92 & 97.11 & 0.261 & 64.60 & 83.68 & 92.18 & 93.70 & 97.79 & 91.84 & 0.489 & 77.32 \\
AE4AD \cite{cai2024rethinking} & 56.96 & 61.07 & 34.82 & 49.49 & 50.59 & 0.073 & 45.94 & 54.29 & 61.06 & 33.86 & 54.16 & 50.84 & 0.037 & 47.93 \\
MMAD-4o & 98.00 & 100.0 & 100.0 & 100.0 & 99.50 & 0.938 & 95.89 & 80.56 & 99.97 & 100.0 & 100.0 & 95.13 & 0.908 & 94.90 \\
MMAD-4o-mini & 99.26 & 99.40 & 99.40 & 99.40 & 99.37 & 0.622 & 71.40 & 75.64 & 98.80 & 98.80 & 98.80 & 93.01 & 0.745 & 81.00 \\
MMAD-Sonnet & 99.36 & 99.99 & 100.0 & 100.0 & 99.84 & 0.962 & 98.29 & 74.53 & 97.28 & 100.0 & 100.0 & 92.95 & 0.911 & 96.34 \\
MMAD-Haiku & 94.64 & 99.77 & 99.88 & 98.14 & 98.11 & 0.790 & 89.58 & 66.04 & 93.43 & 97.52 & 93.34 & 87.58 & 0.738 & 87.15 \\
\hline
        & \multicolumn{7}{c|}{\textbf{golden\_retriever}} 
        & \multicolumn{7}{c}{\textbf{labrador\_retriever}}\\
\cline{1-15}
Skip-GAN \cite{akccay2019skip} & 77.63 & 88.52 & 84.17 & 99.35 & 87.42 & 0.526 & 79.41 & 66.81 & 72.34 & 84.30 & 98.56 & 80.50 & 0.529 & 79.54 \\
RD4AD  \cite{deng2022anomaly} & 48.12 & 63.34 & 72.42 & 80.57 & 66.11 & 0.331 & 68.50 & 44.67 & 59.42 & 67.04 & 74.60 & 61.43 & 0.281 & 65.69 \\
PatchCore \cite{roth2022towards} & 58.26 & 79.66 & 78.19 & 89.99 & 76.53 & 0.429 & 73.95 & 55.54 & 78.62 & 75.18 & 88.30 & 74.41 & 0.409 & 72.84 \\
CFLOW-AD \cite{gudovskiy2022cflow} & 77.03 & 94.75 & 95.91 & 95.43 & 90.78 & 0.491 & 77.44 & 69.52 & 92.76 & 92.40 & 93.10 & 86.94 & 0.468 & 76.15 \\
RRD \cite{tien2023revisiting} & 69.13 & 82.34 & 89.01 & 97.19 & 84.42 & 0.538 & 80.05 & 55.54 & 71.06 & 80.96 & 93.14 & 75.17 & 0.472 & 76.40 \\
OCR-GAN \cite{liang2023omni} & 69.46 & 72.48 & 61.81 & 93.75 & 74.38 & 0.342 & 69.12 & 69.33 & 69.22 & 55.51 & 90.52 & 71.14 & 0.278 & 65.54 \\
PNI \cite{bae2023pni} & 60.61 & 78.41 & 75.63 & 89.51 & 76.04 & 0.429 & 73.99 & 58.47 & 79.05 & 73.92 & 89.60 & 75.26 & 0.414 & 73.15 \\
SPR \cite{shin2023anomaly} & 54.94 & 50.44 & 77.32 & 78.32 & 65.26 & 0.297 & 66.60 & 48.21 & 45.86 & 68.97 & 70.89 & 58.48 & 0.241 & 63.44 \\
IGD \cite{chen2022deep} & 89.61 & 95.63 & 95.35 & 97.30 & 94.48 & 0.468 & 76.13 & 90.53 & 96.75 & 96.64 & 97.20 & 95.28 & 0.436 & 74.38 \\
AE4AD \cite{cai2024rethinking} & 55.09 & 60.99 & 33.96 & 51.96 & 50.50 & 0.052 & 47.09 & 47.22 & 54.46 & 29.29 & 45.52 & 44.12 & 0.093 & 44.83 \\
MMAD-4o & 98.80 & 100.0 & 100.0 & 100.0 & 99.70 & 0.949 & 96.62 & 96.56 & 100.0 & 100.0 & 100.0 & 99.14 & 0.936 & 95.43 \\
MMAD-4o-mini & 99.22 & 100.0 & 100.0 & 100.0 & 99.80 & 0.781 & 82.73 & 98.26 & 100.0 & 100.0 & 100.0 & 99.56 & 0.786 & 83.51 \\
MMAD-Sonnet & 97.99 & 99.89 & 99.89 & 100.0 & 99.44 & 0.952 & 98.17 & 92.06 & 99.82 & 100.0 & 100.0 & 97.97 & 0.957 & 98.74 \\
MMAD-Haiku & 90.93 & 98.35 & 99.68 & 99.17 & 97.03 & 0.790 & 90.06 & 78.53 & 94.48 & 97.54 & 96.36 & 91.73 & 0.777 & 89.71 \\
\hline
        & \multicolumn{7}{c|}{\textbf{teddy}} 
        & \multicolumn{7}{c}{\textbf{mean}}\\
\cline{1-15}
Skip-GAN \cite{akccay2019skip} & 76.06 & 79.47 & 90.15 & 99.03 & 86.18 & 0.564 & 81.53 & 78.62 & 85.63 & 85.93 & 99.31 & 87.37 & 0.541 & 80.25 \\
RD4AD  \cite{deng2022anomaly} & 54.06 & 66.10 & 74.43 & 79.77 & 68.59 & 0.326 & 68.23 & 53.14 & 64.86 & 72.71 & 79.07 & 67.44 & 0.308 & 67.20 \\
PatchCore \cite{roth2022towards} & 58.63 & 78.79 & 77.25 & 88.94 & 75.90 & 0.411 & 72.96 & 58.60 & 78.55 & 76.91 & 88.77 & 75.71 & 0.410 & 72.94 \\
CFLOW-AD \cite{gudovskiy2022cflow} & 82.99 & 94.74 & 94.88 & 92.81 & 91.35 & 0.394 & 72.00 & 76.97 & 93.24 & 93.96 & 93.63 & 89.45 & 0.431 & 74.11 \\
RRD \cite{tien2023revisiting} & 64.18 & 77.07 & 86.42 & 95.41 & 80.77 & 0.509 & 78.43 & 66.14 & 78.26 & 86.40 & 95.87 & 81.67 & 0.510 & 78.51 \\
OCR-GAN \cite{liang2023omni} & 74.02 & 69.05 & 68.37 & 94.94 & 76.60 & 0.375 & 70.97 & 73.04 & 71.78 & 63.73 & 93.96 & 75.63 & 0.343 & 69.19 \\
PNI \cite{bae2023pni} & 63.90 & 79.67 & 75.30 & 90.08 & 77.24 & 0.409 & 72.87 & 63.75 & 79.39 & 75.83 & 89.80 & 77.20 & 0.413 & 73.11 \\
SPR \cite{shin2023anomaly} & 59.74 & 51.80 & 78.34 & 65.12 & 63.75 & 0.188 & 60.48 & 56.88 & 51.62 & 76.97 & 71.70 & 64.29 & 0.242 & 63.55 \\
IGD \cite{chen2022deep} & 92.81 & 97.26 & 97.08 & 98.44 & 96.40 & 0.464 & 75.92 & 90.86 & 95.84 & 95.84 & 97.53 & 95.02 & 0.424 & 73.67 \\
AE4AD \cite{cai2024rethinking} & 48.71 & 57.79 & 31.07 & 49.07 & 46.66 & 0.065 & 46.38 & 52.45 & 59.07 & 32.60 & 50.04 & 48.54 & 0.064 & 46.43 \\
MMAD-4o & 95.08 & 99.90 & 99.99 & 100.0 & 98.74 & 0.934 & 96.72 & 93.80 & 99.97 & 100.0 & 100.0 & 98.44 & 0.933 & 95.91 \\
MMAD-4o-mini & 98.35 & 100.0 & 100.0 & 100.0 & 99.59 & 0.780 & 83.12 & 94.15 & 99.64 & 99.64 & 99.64 & 98.27 & 0.743 & 80.35 \\
MMAD-Sonnet & 97.28 & 99.85 & 99.95 & 100.0 & 99.27 & 0.899 & 95.15 & 92.24 & 99.37 & 99.97 & 100.0 & 97.89 & 0.936 & 97.34 \\
MMAD-Haiku & 75.13 & 94.22 & 97.88 & 95.59 & 90.70 & 0.739 & 86.27 & 81.05 & 96.05 & 98.50 & 96.52 & 93.03 & 0.766 & 88.55 \\
    \bottomrule
    
    \end{tabular}}
    \caption{Full results on MultiDogs-MAD}
    \label{tab:multidogs_detailed}
\end{table*}

%% file: sec/tables/diabetic_detailed.tex
\begin{table*}[htbp]\scriptsize
    \centering
    \renewcommand{\arraystretch}{1.2}
    \scalebox{1}{
\setlength{\tabcolsep}{2.85pt} % Adjust column separation
    \begin{tabular}{p{2cm}ccccccc}
    \toprule
        \multirow{3}{*}{\textbf{Method}} 
        & \multicolumn{5}{c}{\textbf{Binary AD performance}} 
        & \multicolumn{2}{c}{\textbf{MAD performance}}\\
\cline{2-8}

        & \textbf{Level 1} & \textbf{Level 2} & \textbf{ Level 3} & \textbf{ Level 4}
        & \textbf{Whole} & \textbf{\;Ken} & \textbf{C} \\

\cline{1-8}
    % Example row, replace with actual data
    % \hline
Skip-GAN \cite{akccay2019skip} & 52.50 & 49.33 & 51.03 & 52.61 & 51.36 & 0.014 & 50.77 \\
RD4AD  \cite{deng2022anomaly} & 51.96 & 55.12 & 66.17 & 71.94 & 61.30 & 0.217 & 62.14 \\
PatchCore \cite{roth2022towards} & 50.05 & 54.15 & 63.27 & 77.95 & 61.36 & 0.259 & 64.47 \\
CFLOW-AD \cite{gudovskiy2022cflow} & 50.14 & 52.10 & 61.23 & 77.37 & 60.21 & 0.238 & 63.30 \\
RRD \cite{tien2023revisiting} & 50.86 & 55.84 & 65.77 & 74.84 & 61.83 & 0.243 & 63.55 \\
OCR-GAN \cite{liang2023omni} & 52.51 & 55.20 & 47.89 & 59.72 & 53.83 & 0.054 & 53.03 \\
PNI \cite{bae2023pni} & 48.09 & 55.60 & 66.67 & 79.66 & 62.50 & 0.285 & 65.94 \\
SPR \cite{shin2023anomaly} & 47.09 & 47.15 & 44.03 & 47.02 & 46.32 & 0.034 & 48.11 \\
IGD \cite{chen2022deep} & 45.65 & 48.57 & 52.24 & 70.87 & 54.33 & 0.170 & 59.51 \\
AE4AD \cite{cai2024rethinking} & 48.56 & 48.17 & 46.49 & 55.42 & 49.66 & 0.029 & 51.63 \\
MMAD-4o & 49.83 & 59.86 & 75.54 & 77.98 & 65.80 & 0.433 & 67.72 \\
MMAD-4o-mini & 49.98 & 56.50 & 66.87 & 80.86 & 63.55 & 0.348 & 66.66 \\
MMAD-Sonnet & 47.39 & 58.13 & 70.85 & 83.70 & 65.02 & 0.403 & 69.30 \\
MMAD-Haiku & 47.66 & 48.95 & 55.20 & 61.74 & 53.39 & 0.125 & 56.20 \\
    \bottomrule
    
    \end{tabular}}
    \caption{Full results on DRD-MAD}
    \label{tab:diabetic_detailed}
\end{table*}

%% file: sec/tables/covid19_detailed.tex
\begin{table*}[htbp]\scriptsize
    \centering
    \renewcommand{\arraystretch}{1.2}
    \scalebox{1}{
\setlength{\tabcolsep}{2.85pt} % Adjust column separation
    \begin{tabular}{p{2cm}ccccccccc}
    \toprule
        \multirow{3}{*}{\textbf{Method}} 
        & \multicolumn{7}{c}{\textbf{Binary AD performance}} 
        & \multicolumn{2}{c}{\textbf{MAD performance}}\\
\cline{2-10}

        & \textbf{Level 1} & \textbf{Level 2} & \textbf{ Level 3} & \textbf{ Level 4} & \textbf{ Level 5} & \textbf{ Level 6}
        & \textbf{Whole} & \textbf{\;Ken} & \textbf{C} \\

\cline{1-10}
    % Example row, replace with actual data
    % \hline
Skip-GAN \cite{akccay2019skip} & 68.58 & 64.48 & 83.27 & 91.49 & 55.52 & 61.28 & 68.50 & 0.006 & 50.34 \\
RD4AD  \cite{deng2022anomaly} & 79.79 & 74.50 & 85.02 & 81.76 & 86.21 & 87.70 & 83.48 & 0.219 & 61.90 \\
PatchCore \cite{roth2022towards} & 72.21 & 71.98 & 74.52 & 74.56 & 77.75 & 81.67 & 76.33 & 0.200 & 60.88 \\
CFLOW-AD \cite{gudovskiy2022cflow} & 68.97 & 66.65 & 75.52 & 72.39 & 76.27 & 78.99 & 74.03 & 0.180 & 59.78 \\
RRD \cite{tien2023revisiting} & 79.35 & 79.00 & 80.85 & 84.08 & 86.74 & 89.60 & 84.36 & 0.240 & 63.07 \\
OCR-GAN \cite{liang2023omni} & 78.63 & 41.79 & 79.62 & 63.28 & 60.57 & 66.43 & 65.46 & 0.038 & 52.09 \\
PNI \cite{bae2023pni} & 85.23 & 83.19 & 87.02 & 84.81 & 90.29 & 92.12 & 87.96 & 0.241 & 63.12 \\
SPR \cite{shin2023anomaly} & 54.64 & 41.92 & 56.87 & 55.69 & 59.63 & 55.69 & 55.14 & 0.065 & 53.51 \\
IGD \cite{chen2022deep} & 78.75 & 82.62 & 80.87 & 82.65 & 89.20 & 93.11 & 85.82 & 0.312 & 66.99 \\
AE4AD \cite{cai2024rethinking} & 68.85 & 68.75 & 78.15 & 71.24 & 74.75 & 80.97 & 74.45 & 0.189 & 60.26 \\
MMAD-4o & 75.05 & 80.81 & 82.28 & 88.67 & 93.66 & 97.02 & 88.07 & 0.547 & 76.94 \\
MMAD-4o-mini & 66.17 & 80.05 & 75.10 & 81.69 & 82.63 & 86.33 & 79.58 & 0.351 & 66.98 \\
MMAD-Sonnet & 77.46 & 86.60 & 90.58 & 95.78 & 97.11 & 99.10 & 92.35 & 0.601 & 80.54 \\
MMAD-Haiku & 56.65 & 57.01 & 59.54 & 59.60 & 60.44 & 61.21 & 59.42 & 0.142 & 54.16 \\
    \bottomrule
    
    \end{tabular}}
    \caption{Full results on Covid19-MAD}
    \label{tab:covid19_detailed}
\end{table*}

%% file: sec/tables/skin-lesion_detailed.tex
\begin{table*}[htbp]\scriptsize
    \centering
    \renewcommand{\arraystretch}{1.2}
    \scalebox{1}{
\setlength{\tabcolsep}{2.85pt} % Adjust column separation
    \begin{tabular}{p{2cm}cccccc}
    \toprule
        \multirow{3}{*}{\textbf{Method}} 
        & \multicolumn{4}{c}{\textbf{Binary AD performance}} 
        & \multicolumn{2}{c}{\textbf{MAD performance}}\\
\cline{2-7}

        & \textbf{Level 1} & \textbf{Level 2} & \textbf{ Level 3}
        & \textbf{Whole} & \textbf{\;Ken} & \textbf{C} \\

\cline{1-7}
    % Example row, replace with actual data
    % \hline
Skip-GAN \cite{akccay2019skip} & 99.59 & 99.94 & 99.40 & 99.60 & 0.406 & 73.65 \\
RD4AD  \cite{deng2022anomaly} & 98.71 & 98.91 & 99.24 & 98.94 & 0.509 & 79.60 \\
PatchCore \cite{roth2022towards} & 100.0 & 99.99 & 100.0 & 100.0 & 0.456 & 76.57 \\
CFLOW-AD \cite{gudovskiy2022cflow} & 99.99 & 99.94 & 99.99 & 99.98 & 0.389 & 72.63 \\
RRD \cite{tien2023revisiting} & 99.25 & 99.57 & 99.85 & 99.53 & 0.558 & 82.48 \\
OCR-GAN \cite{liang2023omni} & 99.61 & 99.32 & 99.58 & 99.53 & 0.391 & 72.76 \\
PNI \cite{bae2023pni} & 100.0 & 100.0 & 100.0 & 100.0 & 0.455 & 76.51 \\
SPR \cite{shin2023anomaly} & 90.57 & 93.59 & 90.77 & 91.31 & 0.335 & 69.50 \\
IGD \cite{chen2022deep} & 97.64 & 96.71 & 99.03 & 97.91 & 0.487 & 78.34 \\
AE4AD \cite{cai2024rethinking} & 99.02 & 98.37 & 99.79 & 99.13 & 0.505 & 79.42 \\
MMAD-4o & 99.49 & 98.75 & 99.80 & 99.43 & 0.694 & 85.61 \\
MMAD-4o-mini & 99.52 & 99.90 & 99.95 & 99.75 & 0.653 & 84.97 \\
MMAD-Sonnet & 99.79 & 99.80 & 99.87 & 99.82 & 0.610 & 79.23 \\
MMAD-Haiku & 99.99 & 99.67 & 99.78 & 99.85 & 0.479 & 74.19 \\
    \bottomrule
    
    \end{tabular}}
    \caption{Full results on SkinLesion-MAD}
    \label{tab:skin-lesion_detailed}
\end{table*}